\documentclass{article}
\usepackage[T1]{fontenc}

\usepackage{microtype}
\usepackage{graphicx}
\usepackage{subcaption}
\usepackage{booktabs} 
\usepackage{xurl}
\usepackage{tabularx} 

\usepackage{amssymb}
\usepackage{pifont}
\newcommand{\cmark}{\ding{51}}%
\newcommand{\xmark}{\ding{55}}%

\usepackage{hyperref}

\usepackage[table]{xcolor}

\definecolor{Gray}{gray}{0.9}

\usepackage[accepted]{icml2026}

\usepackage{amsmath}
\usepackage{amssymb}
\usepackage{mathtools}
\usepackage{amsthm}
\usepackage{breqn}
\usepackage{xspace}
\usepackage{subcaption}
\usepackage{siunitx}
\usepackage{makecell}
\usepackage{wrapfig}
\usepackage{pifont}
\usepackage{placeins}
\usepackage[normalem]{ulem} 

\usepackage[capitalize,noabbrev]{cleveref}

\theoremstyle{plain}

\theoremstyle{definition}

\theoremstyle{remark}

\newcolumntype{Y}{>{\centering\arraybackslash}X}

\newcommand{\ours}{SynLaD\xspace}

\icmltitlerunning{\ours: Latent Diffusion for Generating Synthesizable Molecules Conditioned on 3D Pharmacophore Profiles}

\begin{document}

\twocolumn[
  \icmltitle{\ours: Latent Diffusion for Generating Synthesizable Molecules Conditioned on 3D Pharmacophore Profiles}



  \icmlsetsymbol{workdone}{*}

  \begin{icmlauthorlist}
    \icmlauthor{Miruna Cretu}{cam,genentech,workdone}
    \icmlauthor{John Bradshaw}{genentech}
    \icmlauthor{Patricia Suriana}{genentech}
    \icmlauthor{Saeed Saremi}{genentech}
    \icmlauthor{Omar Mahmood}{genentech}
    \icmlauthor{Kirill Shmilovich}{genentech}
    \icmlauthor{Kangway Chuang}{genentech}
    \icmlauthor{Vishnu Sresht}{genentech}
    \icmlauthor{Colin Grambow}{genentech}
  \end{icmlauthorlist}

  \icmlaffiliation{cam}{University of Cambridge, Cambridge, UK}
  \icmlaffiliation{genentech}{Prescient Design (AI for Drug Discovery), Genentech, South San Francisco, USA}

  \icmlcorrespondingauthor{Miruna Cretu}{mtc49@cam.ac.uk}
  \icmlcorrespondingauthor{John Bradshaw}{bradshaw.john@gene.com}
  \icmlcorrespondingauthor{Colin Grambow}{grambow.colin@gene.com}

  \icmlkeywords{Machine Learning, ICML}

  \vskip 0.3in
]



\printAffiliationsAndNotice{\textsuperscript{*}Work done during an internship at Prescient Design.}

\begin{abstract}

We present \ours, a latent diffusion framework for small-molecule generation that unifies ligand-based drug design objectives (what to make) with synthetic accessibility (how to make it). Current models typically optimize one objective at the expense of the other, creating a bottleneck for discovering high-scoring and synthesizable molecules. \ours combines reaction-constrained generation with pharmacophore-conditioned 3D design by learning a latent space that decodes to both 3D structures and synthesis pathways. An encoder maps molecules to a latent representation used by two decoder heads: (i) a geometric head that reconstructs atom types and coordinates and (ii) an autoregressive synthesis head that outputs synthetic routes in a serialized, reaction-based notation. A diffusion transformer generates novel latents in the learned space, conditioned on pharmacophore profiles. Across analogue generation tasks for bioactive ligands, \ours outperforms existing baselines in synthesizable and diverse \textit{hit} generation, demonstrating that a single model can produce shape-aligned molecules with feasible synthesis plans.

\end{abstract}

\section{Introduction}

Ligand-based drug design (LBDD) uses known bioactive compounds to guide the design of new molecules with similar three-dimensional (3D) shapes and physicochemical properties \citep{acharya2011recent}. In contrast to structure-based drug design (SBDD), which relies on accurate protein structures and is often more computationally demanding, LBDD remains effective when target structures are unavailable and is therefore widely used for hit discovery and hit diversification \citep{Grebner2020}. Central to LBDD is the pharmacophore, which specifies the spatial arrangement of interaction features—such as hydrogen bond donors or acceptors, charged groups, and aromatic centers—responsible for protein-ligand complementarity.

A common LBDD strategy extracts pharmacophores from known ligands and uses them in virtual screening to identify molecules with similar shapes and features \citep{goodnow2007hit}. Methods such as ROCS follow this approach and have shown competitive or better performance than structure-based methods \citep{rocs_web, rocs_eg, rocs_eg2}. However, the rapid expansion of accessible chemical space \citep{chem_space_growth} renders exhaustive screening increasingly computationally prohibitive. Generative modeling has emerged as a promising alternative to brute-force search, allowing for faster inference times and the proposing of molecules outside of predefined spaces. Conditional generative models for 3D small molecule design span a range of families, from autoregressive \citep{peng2025pocket2molefficientmolecularsampling} to diffusion \citep{luo20223dgenerativemodelstructurebased, schneuing2024structurebaseddrugdesignequivariant, guan20233dequivariantdiffusiontargetaware, pilot}, flow matching \citep{schneuing2025multidomain, cremer2025flowrflowmatchingstructureaware}, and variational autoencoders \citep{zhu2023pharmacophore}. While performing well at capturing distributions of desired molecular motifs, these models often produce synthetically inaccessible molecules \citep{gao_synth, STANLEY2023102658}.

\begin{figure*}[t] 
    \centering
    \includegraphics[width=\textwidth]{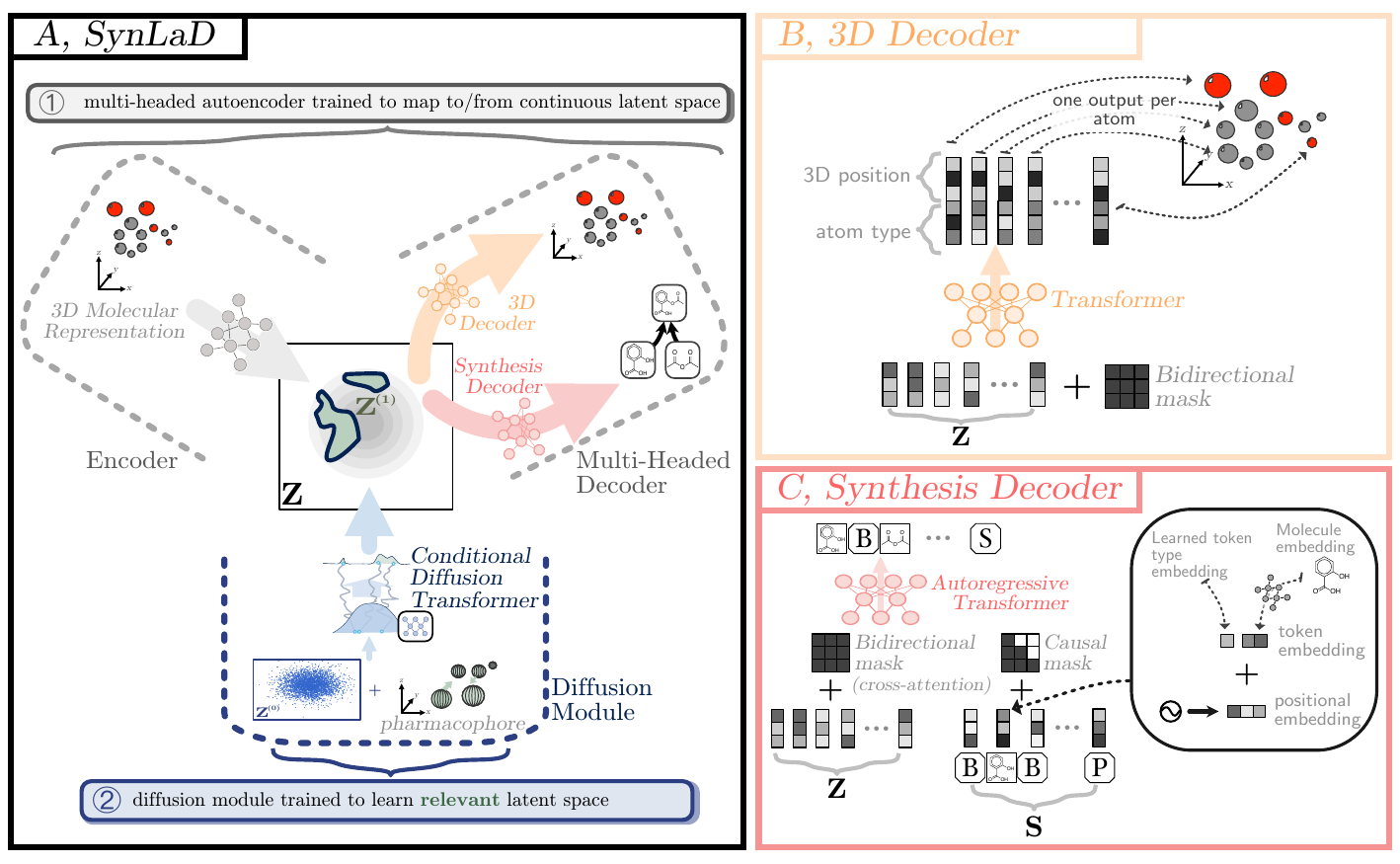}
    \caption{
    \textbf{A, Overview of \ours: a multi-head generative model that jointly produces 3D molecular structures and synthesis plans.}
    \ours is trained in two stages: \textbf{\textcolor[HTML]{595959}{\ding{172}}} an autoencoder learns a latent representation of 3D molecules, with an auxiliary decoder that generates a synthesis plan for each structure.
    \textbf{\textcolor[HTML]{313B80}{\ding{173}}} a pharmacophore-conditioned diffusion model learns to sample latents that decode to molecules containing a desired pharmacophore. 
    \textbf{B, 3D Decoder:} a transformer with a bidirectional mask maps latent representations to atom types and 3D coordinates.
    \textbf{C, Synthesis Decoder:} an autoregressive transformer maps latent representations to serialized synthesis plans. 
    At inference time, the synthesis decoder is conditioned on the 3D-informed latent via cross-attention, enabling \ours to generate outputs that inherit pharmacophore-aware structural information while remaining directly tied to a synthesizable route.
    }
    \vspace{-10pt}
    \label{fig:modelOverview}
\end{figure*}

Synthesis-aware generative methods for designing new molecules have typically either treated synthesizability as an additional property to optimize \citep{liu2022retrognn-714,guo2025directly-aea,gomezbombarelli2018automatic-f5e} or enforced it
 by generating molecules via synthetic pathways/reactions rather than as unconstrained molecular graphs \citep{hartenfeller2012dogs-a10,Vinkers2003,noutahi_synth, NEURIPS2019_46d0671d, gao2022amortizedtreegenerationbottomup, swanson2024generative, cretu2025synflownetdesigndiversenovel}. However, these approaches are not conditioned on three-dimensional chemical features, which are critical for protein-ligand binding. As a result, bioactive ligand generation is typically framed either as a reward-based problem using binding-affinity proxies \citep{noutahi_synth, cretu2025synflownetdesigndiversenovel} or as conditioning on known ligands via chemical space projection \citep{luo2025projectingmoleculessynthesizablechemical}. 

We bridge the gap between 3D chemical feature conditioning and synthesizable molecule generation with \ours (\textit{\uline{Syn}thesis-aware \uline{La}tent \uline{D}iffusion}), a conditional latent diffusion model that decodes latents to molecules in both 3D and synthesizable space. An overview of \ours is shown in Figure~\ref{fig:modelOverview}. At its core, \ours employs a molecular variational autoencoder with a dual-decoder architecture: a single encoder maps a 3D molecule to the latent space, from which one decoder reconstructs the 3D atomic coordinates and corresponding atom types, while a second autoregressive decoder generates an associated synthesis pathway.  Joint training of these decoders couples spatial fidelity and synthesizability in the learned latent representation. 

Conditional generation is then performed in the learned latent space using a Diffusion Transformer (DiT) \citep{rombach2022highresolutionimagesynthesislatent, joshi2025allatom}. 
The compressed, semantically rich latent space enables efficient sampling and improved conditional control \citep{rombach2022highresolutionimagesynthesislatent, liu2023audioldmtexttoaudiogenerationlatent}, while also allowing \ours to generate structure-informed embeddings that the synthesis head decodes into synthesis pathways. 
This also introduces variability in the synthesis decoder conditioning input, which we demonstrate is essential for ensuring sample diversity.  Contrary to recent synthesis-constrained generative models \citep{swanson2024generative, cretu2025synflownetdesigndiversenovel, koziarski2024rgfn, gao2024generative, lee2025rethinking} which rely on reaction templates, \ours uses a reaction prediction model trained on diverse, complementary chemical transformations to infer products. By removing the reliance on fixed templates, our approach mitigates the template bottleneck that can restrict designs to a narrow, predefined chemical space, and supports broader exploration via more general molecular-graph edits, and the possibility of extrapolation.

We evaluate \ours using standard molecule quality metrics, including pose validity and synthesizability, under both unconditional and pharmacophore-conditioned generation. We further assess its practical utility on bioactive analogue generation for ligands from the Lit-PCBA benchmark \citep{litpcba}. Across settings, \ours matches the molecule-quality profile of its separately trained counterparts while consistently improving the fraction of samples with plausible synthetic routes, and it outperforms existing baselines in producing synthesizable, diverse hits under pharmacophore-guided sampling.

The contributions of this work are thus: (1) a dual-constrained latent space that couples two complementary but often competing representations—3D molecular structure and synthesis plans—encouraging latents to remain simultaneously decodable into valid geometries and executable routes; (2) a pharmacophore-conditional latent generative model that enables targeted, 3D-aware design; and (3) an end-to-end evaluation protocol that jointly reports diversity, synthesizability, and target-conditioned hit discovery trade-offs—explicitly contrasting our conditional generation method with library screening and other amortized and non-amortized generative baselines. We release the code for \ours at \hyperlink{https://github.com/prescient-design/synlad}{https://github.com/prescient-design/synlad}.

\vspace{-8pt}



\section{Background and related work}

\paragraph{Structure- and ligand-based drug design.}
Many generative models have been proposed to sample small molecules in sequence \citep{Gupta2018GenerativeRNN, Blaschke2020,segler2018generating}, 2D \citep{C8SC05372C, jin2017predicting, qin2025defogdiscreteflowmatching}, 3D \citep{le2023navigatingdesignspaceequivariant, dunn2024mixedcontinuouscategoricalflow, huang2025midimultiinstancediffusionsingle, irwin2025semlaflowefficient3d, vonessen2025tabascofastsimplifiedmodel}, and voxel spaces \citep{pinheiro20243dmoleculegenerationdenoising}. Conditional generation for protein binders can be broadly categorized into pocket-conditioned approaches, which condition on binding-site representations and implicitly capture protein-ligand interactions, and interaction-conditioned approaches, which condition directly on desired interaction cues such as 3D shape, pharmacophores, or electrostatic potential surfaces. The latter is most relevant to our work. Several methods generate SMILES strings or molecular graphs conditioned on 3D pharmacophore representations \citep{D1SC02436A, zhu2023pharmacophore, xie2024acceleratingdiscoverynovelbioactive, mahmood2025pharmacophorebaseddesignlearningvoxel}. MolSnapper \citep{ziv_molsnapper} and, more recently, ShEPhERD \citep{adamsShEPhERD2024} instead condition on richer interaction profiles and generate molecules directly in 3D space. However, MolSnapper supports a limited set of pharmacophore types, and neither method explicitly constrains generation to synthesizable chemical space.

\vspace{-13pt}

\paragraph{Synthesis-aware generation.}
Parallel work has focused on enhancing synthesizability by generating molecules via reaction pathways.
Composing these reaction pathways into full synthesis plans is a challenging discrete search problem, and existing methods span diverse paradigms, including reinforcement learning \citep{gottipati2020learning, noutahi_synth}, search \citep{swanson2025synthemol}, surrogate-guided optimization \citep{korovina2020chembo}, genetic algorithms \citep{lo2025genetic}, and GFlowNets \citep{cretu2025synflownetdesigndiversenovel, koziarski2024rgfn,seo2025generativeflowssyntheticpathway}. One particularly effective strategy is to cast synthesis plans as sequences and learn \emph{serialized} pathway representations autoregressively \citep{bradshaw_dog,gao2022amortizedtreegenerationbottomup,lee2025rethinking,luo2025projectingmoleculessynthesizablechemical,gao2024generative}, analogous to next-token prediction in language modeling \citep{NIPS2017_3f5ee243, bengio2003neural, sutskever2014sequence}. By amortizing the cost of discrete search, such models learn mappings from continuous representations to synthesis plans and can be integrated into larger frameworks (e.g., encoder-decoder architectures) for tasks such as molecular optimization or retrosynthesis (i.e., predicting a route for a given target molecule). While these approaches enforce viable synthesis routes, they do not directly control final 3D geometry. \citet{luo2025efficientprogrammableexplorationsynthesizable} recently proposed a framework to train an autoregressive synthesis pathway generator conditioned on composite properties, including pharmacophore-encoding fingerprints \citep{gobbi} and molecular substructures. However, the model does not condition on raw 3D pharmacophore features. Recently, CGFlow \citep{shen2025compositionalflows3dmolecule} and SynCoGen \citep{rekesh2025syncogensynthesizable3dmolecule} jointly design synthesis pathways and 3D poses, but they rely on fixed reaction template libraries and work in ambient space. 

\vspace{-10pt}

\paragraph{Latent diffusion models.}
Latent diffusion models \citep{NEURIPS2021_5dca4c6b, rombach2022highresolutionimagesynthesislatent} perform diffusion \citep{pmlr-v37-sohl-dickstein15,Song2019GenerativeMB} in the learned latent space of an autoencoder rather than directly in the high-dimensional input space, enabling more efficient training and sampling. This paradigm has been highly effective in image, audio, and video generation, especially when combined with Diffusion Transformers (DiTs) \citep{Peebles_2023_ICCV}, which show that standard transformer backbones scale effectively as denoisers. In molecular modeling, \citet{xu2023geometriclatentdiffusionmodels} introduced latent diffusion in the space of an equivariant autoencoder, while \citet{joshi2025allatom} proposed a (non-equivariant) latent diffusion model for small molecules and periodic materials. However, these approaches do not demonstrate latent generation conditioned on rich, chemically grounded features. We address this gap by enabling pharmacophore-conditioned latent generation and augmenting our autoencoder with an auxiliary synthesis-decoder head that produces explicit reaction pathways, jointly targeting native-3D conditional generation and synthesizability.


\section{Methods}

\vspace{-5pt}

We present \textsc{\ours}---a latent diffusion \citep{rombach2022highresolutionimagesynthesislatent} model for \textit{de novo} small molecule generation conditioned on pharmacophore features (see Figure~\ref{fig:modelOverview}). \textsc{\ours} is trained in two stages: In stage 1, a variational autoencoder \citep{kingma2022autoencodingvariationalbayes, rezende2015variational} encodes molecules into a shared latent space and a two-headed decoder reconstructs both the 3D molecular structure and a synthesis pathway. 
In stage 2, we train a Diffusion Transformer \citep{Peebles_2023_ICCV} to generate new samples from the latent space, which are decoded into both 3D molecules and synthesis pathways. The two decoders are trained to reconstruct the \textit{same} molecule under two \textit{different} representations and at inference time the synthesis decoder is used to ensure the synthesizability of generated designs. 

We hypothesize that this two-stage learning process offers two advantages. First, the shared latent space aligns the two modalities---molecular conformations and synthesis pathways---by requiring both decoders to reconstruct the same molecule from a single latent representation. Second, as shown in \citet{joshi2025allatom}, operating in a latent space reduces the complexity of generation and enables explicit pharmacophore conditioning by decoupling discrete and continuous spaces.

\subsection{Multi-headed autoencoder for multi-modal reconstruction}
\label{section:encoder}

Our autoencoder consists of one encoder and two decoders trained using a combined reconstruction loss with a regularization term. We base our implementation of the 3D encoder and decoder on \citet{joshi2025allatom} and represent molecules using discrete atom types $\boldsymbol{A}=\{a_i\}_{i=1}^{N}\in \mathbb{Z}^{1\times N}$ and continuous coordinates $\boldsymbol{X}=\{x_i\}_{i=1}^{N}\in \mathbb{R}^{3\times N}$, where $N$ is the number of atoms. 

\subsubsection{3D Encoder and Decoder} Given a molecule represented as above, the encoder $f$ projects $\boldsymbol{A}$ and $\boldsymbol{X}$ into a per-atom latent representation $\boldsymbol{Z}=f(\boldsymbol{A},\boldsymbol{X})$, and the 3D decoder $g_{\textrm{3D}}$ reconstructs the molecule from the latent, giving $\boldsymbol{A}',\boldsymbol{X}'=g_{\textrm{3D}}(\boldsymbol{Z})$, where $\boldsymbol{Z}=\{z_i\}_{i=1}^{N}\in\mathbb{R}^{d\times N}$. We use a standard transformer \citep{NIPS2017_3f5ee243} and learn molecular symmetries via random rotations; further details on how atoms and dimensions are sampled at training/inference time is provided in Appendix~\ref{sect:pharmCondGen}.

\subsubsection{Synthesis Decoder}
\label{section:methods_synth_decoder}

We define an auxiliary decoder $g_{\textrm{syn}}(\boldsymbol{Z})$, which, given a latent $\boldsymbol{Z}$, generates a synthesis pathway for the molecule represented in 3D by $(\boldsymbol{A}, \boldsymbol{X}) = g_{\textrm{3D}}(\boldsymbol{Z})$.
A synthesis pathway can be represented using a directed acyclic graph (DAG), where nodes represent molecules and edges represent reactions (see Figure~\ref{fig:modelOverview}).
Such a DAG shows how parent building block molecules (source nodes) react to first form more complex intermediate products and eventually a final product (a single sink node).
To model such a DAG we can \emph{serialize} it, in a bottom-up manner similar to \citet{bradshaw_dog}, into a linear sequence of tokens following the recipe in Appendix~\ref{app:networkSerialization}. Once we have such a synthesis sequence $\mathbf{S}=(s_1, \dots, s_L)$, we train a causal transformer model to generate it by autoregressively modeling the conditional probability $p(s_i|\boldsymbol{S_{<i}}, \boldsymbol{Z})$. We embed each token in the sequence using a custom embedding scheme (see Appendix~\ref{app:actionEmbeddings}) and train the transformer using teacher forcing conditioned with cross-attention on the latents.

\vspace{-11pt}

\paragraph{Product identity.} 
In our serialized synthesis plan representation, the identity of each product node is determined by the identities of the corresponding reactant molecules. During training, we use a dataset containing full synthesis plans (see Section~\ref{sect:experiments}), allowing the correct reaction products to be inserted directly. 
At test time, however, product information is not available and must be inferred. We therefore employ a reaction-prediction oracle that can predict the most likely product given a set of reactants. The oracle is a transformer-based encoder-decoder model built off the BART architecture \citep{lewis2020bart}, closely related to the Molecular Transformer of \citet{molecularTransformer}. Details of its training are provided in Appendix~\ref{app:reactionOracle}.

\subsubsection{Training objective} The two decoders are jointly trained with a weighted loss: the 3D decoder uses cross-entropy and MSE losses to reconstruct atom types and coordinates, while the synthesis decoder is optimized with a cross-entropy loss between its output logits and the ground-truth token sequence:

\vspace{-13pt}

\begin{equation}
\label{eq:loss}
\begin{aligned}
\mathcal{L}
&= \lambda \mathcal{L}_{\rm 3D}
 + (1-\lambda)\mathcal{L}_{\text{synthesis}} \\
&\quad + \beta \, D_{\mathrm{KL}}\!\Bigl(
    \mathcal{N}( \mu_{\mathbf{Z}}, \sigma_{\mathbf{Z}})
    \,\Bigm\|\,
    \mathcal{N}(0,I_d)
\Bigr)
\end{aligned}
\end{equation}

\vspace{-15pt}

\noindent where:

\vspace{-18pt}

\begin{align}
    \mathcal{L}_{\rm 3D} &= \frac{1}{N} \sum_{i=1}^{N} H(a_i, \hat{a}_i) + \frac{1}{3N} \sum_{i=1}^{N} \frac{\|\tilde{x}_i - \hat{\tilde{x}}_i\|^2}{\sigma^2}, \\
    \mathcal{L}_{\text{synthesis}} &= \frac{1}{L} \sum_{i=1}^{L} H(s_i, \hat{s}_i).
\end{align}

\vspace{-12pt}

Here, \noindent $\tilde{x}$ represents zero-centered coordinates, $\sigma^2$ is $1.0$, $0\leq \lambda \leq 1$ is the linear interpolation parameter between $\mathcal{L}_{\text{synthesis}}$ and $\mathcal{L}_{\rm 3D}$, $L$ is the total sequence length, and $s_i$ is the ground-truth synthesis token at step $i$. Following \citet{Higgins2016betaVAELB}, \citet{rombach2022highresolutionimagesynthesislatent}, and \citet{joshi2025allatom} we introduce a per-channel KL-penalty weighted by $\beta$ that regularizes the learned latent toward a standard normal distribution.

\subsection{Conditional generative modeling of latent representations}

\label{section:conditioning_method}
\vspace{-2pt}
We train a flow matching model \citep{Lipman2022FlowMF, Song2019GenerativeMB, Ho2020DenoisingDP,joshi2025allatom} on the latent space learned by our autoencoder, by denoising latent samples from a base Gaussian distribution towards a target distribution represented by latents of ground-truth molecules.

\vspace{-10pt}

\paragraph{Pharmacophore embeddings.} A pharmacophore abstracts components of a molecule into pre-defined key interaction features, providing a compact representation of interaction-relevant moieties and their spatial coordinates. We represent pharmacophores using six discrete pharmacophore types $\boldsymbol{P}=\{p_i\}_{i=1}^{N_p}\in \mathbb{Z}^{1\times N_p}$ (hydrogen bond donor, hydrogen bond acceptor, cation, anion, aromatic ring and hydrophobe) and coordinates for each pharmacophore feature present in the molecule: $\boldsymbol{X^{\text{ph}}}=\{x^{\text{ph}}_{i}\}_{i=1}^{N_p}\in \mathbb{R}^{3\times N_p}$. These are extracted from the molecule conformation through a deterministic mapping of the pharmacophore's atom positions. The features are embedded into a unified learned representation using a standard transformer (see Appendix \ref{app:hyperparam}), and this embedding serves as conditioning for the diffusion model via cross-attention. The parameters of the pharmacophore transformer encoder and those of the DiT are jointly optimized via the conditional flow matching loss:

\vspace{-20pt}

{
\setlength{\jot}{-2pt}

\begin{multline}
\mathcal{L}_{\text{CFM}}
= \mathbb{E}_{\substack{
(\boldsymbol{P}, \boldsymbol{X}^{\text{ph}}, \boldsymbol{Z}_1)\sim p_{\text{data}},\\
t \sim \mathcal{U}(0,1),\;
\boldsymbol{Z}_0 \sim \mathcal{N}(0,I)
}}
\Big[
    \big\|
        u_\theta\!\big(
            \boldsymbol{Z}_t,\; t,\; \tau_\theta(\boldsymbol{P}, \boldsymbol{X}^{\text{ph}})
        \big) 
       \\ - v_t\!\big(\boldsymbol{Z}_t \mid \boldsymbol{Z}_0,\boldsymbol{Z}_1\big)
    \big\|_2^2
\Big], 
\end{multline}
}

\vspace{-10pt}

\noindent where $u_\theta$ is the denoiser backbone, which predicts the vector field based on the current state $z_t$, the time step $t$, and the pharmacophore conditional embedding $\tau_\theta(\boldsymbol{P}, \boldsymbol{X^{\text{ph}}})$. $v_t$ is the ground-truth vector field and the state $\boldsymbol{Z_t} = (1-t)\boldsymbol{Z_0} + t\boldsymbol{Z_1}$ is a linear interpolation between a clean latent sample $\boldsymbol{Z_1}$ and a noise sample $\boldsymbol{Z_0}$ drawn from a standard normal distribution $\mathcal{N}(0,I)$. The time step $t$ is sampled uniformly from $\mathcal{U}(0,1)$. Leveraging the equivalence between the velocity and endpoint flow matching formulations \citep{Lipman2022FlowMF}, the model receives $\boldsymbol{Z_t}$ as input and predicts the terminal points of the trajectory during training. 

Following \citet{joshi2025allatom}, we use a Diffusion Transformer (DiT) \citep{Peebles_2023_ICCV} as our denoiser architecture with self-conditioning \citep{jason_yim_selfcond} and adaptive layer norm for time-step $t$ modulation. We additionally introduce a cross-attention conditioning mechanism (trained with a conditioning dropout probability of 20\%) to sample using classifier-free guidance during inference \citep{ho2022classifierfreediffusionguidance}:

\vspace{-25pt}

\begin{multline}
\label{eq:cfg} 
\hat{u}_\theta(\boldsymbol{Z_t}, t, \tau_\theta(\boldsymbol{P}, \boldsymbol{X^{\text{ph}}})) = (1+w)u_\theta(\boldsymbol{Z_t}, t, \tau_\theta(\boldsymbol{P}, \boldsymbol{X^{\text{ph}}})) \\ - w u_\theta(\boldsymbol{Z_t}, t, \emptyset) 
\end{multline}

\vspace{-12pt}

\noindent where $\hat{u}_\theta$ is the final vector field prediction. This is a linear combination of the conditional and unconditional predictions, where $\emptyset$ represents the null token used for the dropped-out condition. The guidance scale $w$ controls the strength of the conditioning signal and we experiment with both $w=0$ and $w>0$.

\vspace{-6pt}

\section{Experiments}
\label{sect:experiments}

\paragraph{Dataset.} We train \ours on a set of reaction pathways extracted from the USPTO dataset \citep{Lowe2017}. We detail the construction of the dataset in Appendix~\ref{app:dataset}. Our dataset contains 67,512 synthesis pathways, each with 1 to 6 intermediary reactions. For each final product of the pathways, we compute 10 low-energy conformers using OpenEye's conformer generation software Omega \citep{omega, omega_quality}. Our dataset contains synthesis products with 10 to 40 heavy atoms.

\vspace{-8pt}

\paragraph{Metrics.} We evaluate \ours on its ability to generate valid, novel, and realistic molecules. Sampled molecules generated by \ours are assessed using validity and internal diversity \citep{moses}
(definitions provided in Appendix~\ref{app:metrics}). Diversity is further measured by the number of unique Bemis-Murcko scaffolds \citep{bemis_murcko}, defined as the molecule’s ring systems and connecting linkers with all side chains removed. Synthesizability is  additionally evaluated using AiZynthFinder \citep{aizynth}, and pose quality is assessed with the PoseBusters benchmark \citep{buttenschoen2024posebusters} for molecules generated by our 3D decoder. For pharmacophore-conditioned experiments, we use OpenEye's ROCS scoring functions \citep{rocs_eg} to evaluate shape and pharmacophore overlap via Tanimoto Shape and Tanimoto Color scores, each ranging from 0 to 1 (with 0 representing no overlap and 1 representing perfect overlap), whose sum defines the Tanimoto Combo score. Following \citet{Grebner2020}, we define a \textit{hit} as a molecule with a Tanimoto Combo score of at least 1.2 relative to a query. 

As described in Section~\ref{section:methods_synth_decoder}, we use a reaction-prediction oracle to infer product information from sets of reactant tokens sampled by our decoder. The oracle is trained separately; implementation details and accuracy results are provided in Appendix~\ref{app:reactionOracle}, where it achieves a top-1 product prediction accuracy of 84.6\%.

\subsection{Joint conformer and synthesis latent space}
\label{exp:vae_section}

We first investigate reconstruction performance of the joint autoencoder. We randomly select \num{1000} molecules from the test set, encode them into our latent space, and decode each latent using both decoders. For the 3D decoder,  we report match rate and RMSD between ground-truth and predicted coordinates (see Appendix~\ref{app:metrics}); for the synthesis decoder, we report match rate and 2D fingerprint Tanimoto similarity between the predicted end product and the ground-truth synthesis outcome. We ablate the latent dimension during training and compare inference strategies for the synthesis decoder (see Appendix~\ref{app:vae_training}). Results are summarized in Table~\ref{tab:autoencoder_reconstruct} for both \textit{sampling} and \textit{beam search} decoding. \textit{Sampling} generates $N$ independent candidate sequences by temperature-scaled stochastic sampling at each step, optionally using \textit{top-k} truncation by setting logits of tokens outside of the top $k$ to $-\infty$. In contrast, \textit{beam search} deterministically expands the $b$ (number of beams) most probable partial sequences. In our experiments, we set $N=1$ and $k=10$ for sampling and compare against beam search with $b=5$. Beam search consistently outperforms top-$k$ sampling and is therefore used for latent-space sampling in subsequent experiments.

\vspace{-5pt}

\begin{table}[tbh]
  \centering
  \scriptsize
  \setlength{\tabcolsep}{2.5pt}
  \renewcommand{\arraystretch}{1.2}

  \caption{\textbf{Autoencoder reconstruction accuracies.} Results are reported for two synthesis-decoder inference strategies, with synthesis match rate and RDKit fingerprint Tanimoto similarity computed between the final products of predicted and ground-truth synthesis sequences.}
  \label{tab:autoencoder_reconstruct}

  \begin{tabularx}{\columnwidth}{>{\raggedright\arraybackslash}XYYYY}
    \toprule
    \multicolumn{1}{c}{} &
    \multicolumn{2}{c}{\textbf{3D metrics}} &
    \multicolumn{2}{c}{\textbf{Synthesis metrics}} \\
    \cmidrule(r){1-1} \cmidrule(r){2-3} \cmidrule(l){4-5}

    {Inference method} &
    {RMSD (\AA)}$\downarrow$ &
    {Match Rate (\%)}$\uparrow$ &
    {Match Rate (\%)}$\uparrow$ &
    {Tanimoto similarity}$\uparrow$ \\
    \midrule
    Sampling &  0.05 & 98.5 & 43.4 & 0.73 \\
    Beam search & 0.05 & 98.5 & 63.4 & 0.84 \\
    \bottomrule
  \end{tabularx}
\end{table}

\vspace{-5pt}

\subsection{Unconditional generation}

\begin{table*}[h]
  \centering
  \footnotesize 
  
  \caption{\textbf{Unconditional generation results.} $\uparrow$/$\downarrow$ indicate that higher/lower is better. All metrics (see Appendix~\ref{app:metrics}) are evaluated over \num{1000} samples, except for AiZynthFinder (100 samples).}
  \label{tab:uncond_results}
  
  \begin{tabularx}{\textwidth}{ll *{6}{Y}} 
    \toprule
    & 
    \textbf{Method} &
    \textbf{Valid.}$\uparrow$ & 
    \textbf{IntDiv.}$\uparrow$ & 
    \textbf{Nov.}$\uparrow$ & 
    \textbf{PB}$\uparrow$ & 
    \textbf{SA}$\downarrow$ & 
    \textbf{AiZynth.}$\uparrow$  \\
    \midrule
    \textit{\footnotesize{separately-trained heads}}  &  3D Decoder     & 88.5 & 86.0 & 100 & 0.89 & 2.78 & 0.59  \\
    & Synthesis Decoder      & 100 & 89.8 & 87.6 & - & 2.35 & 0.81  \\
    \midrule
    \rowcolor{Gray}
    \textit{\footnotesize{jointly-trained heads}}  &  \ours (3D outputs)        & 90.0 & 86.0 & 100 & 0.89 & 2.65 & 0.72 \\
    \rowcolor{Gray}
    & \ours (synthesis outputs) & 100 & 88.2 & 84.6 & - & 2.30 & 0.87 \\
    \bottomrule
  \end{tabularx}
  \vspace{-5pt}
\end{table*}

We next train an unconditional latent DiT denoiser using latents produced by our joint VAE (variational autoencoder) described in Section~\ref{exp:vae_section}; model parameters and training details are provided in Appendix~\ref{app:dit_training}. We sample using $T=100$ ODE integration steps and decode the resulting latent with both decoders. Results are reported in Table \ref{tab:uncond_results}, where we compare against (1) a latent diffusion model trained without a synthesis decoder head (equivalent to \citet{joshi2025allatom}) and (2) an unconditional autoregressive synthesis-decoder-only head. The results show that jointly training the two decoders does not degrade synthesizability or PoseBusters performance, and yields good validity, diversity, and novelty overall. Interestingly, joint training improves the synthesizability of molecules decoded by the 3D decoder, suggesting that the synthesis decoder steers generation toward latent regions more likely to decode to synthesizable molecules. Both \ours decoders significantly outperform the synthesis-unconstrained baseline in synthesizability although trained on the same data.

\vspace{-9pt}

\begin{figure}[!b]
    \centering
    \vspace{-15pt}
    \begin{minipage}[c]{0.45\columnwidth}
        \centering
        \scriptsize
        \setlength{\tabcolsep}{4pt}
        \renewcommand{\arraystretch}{1.1}
        \resizebox{\linewidth}{!}{%
        \begin{tabular}{@{}lr@{}}
            \toprule
            \textbf{Metric} & \textbf{Value} \\
            \midrule
            Exact match rate (\%) & 33.1 \\
            Scaffold match rate (\%) & 54.9 \\
            \bottomrule
        \end{tabular}
        }
    \end{minipage}
    \hfill
    \begin{minipage}[c]{0.54\columnwidth}
        \centering
        \includegraphics[width=\linewidth]{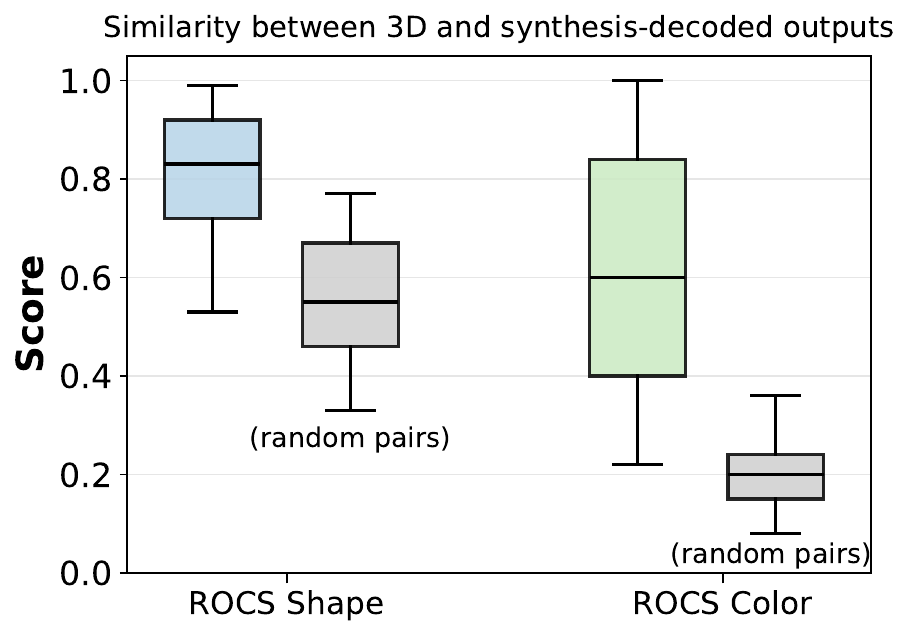}
    \end{minipage}

    \caption{\textbf{Cross-decoder agreement on generated samples.}
    We evaluate similarities between synthesis- and 3D-decoded molecules obtained from the same latent.
    As a control, we also report ROCS Tanimoto Shape/Color scores for randomly paired synthesis- and 3D-decoded molecules,
    illustrating the similarity expected in the absence of cross-decoder coupling.}
    \label{fig:cross_decoder_agreement}
\end{figure}

\paragraph{Consistency analysis.} To enable pharmacophore-conditioned sampling of synthesizable molecules that also match a desired 3D profile, we evaluate how well the two decoders remain aligned on \emph{generated} samples. Starting from a latent $\boldsymbol{Z}$ sampled from the trained diffusion model, we decode both a 3D molecule and a synthesis pathway and quantify how often these outputs correspond to the same, or closely matching, chemical features. We report cross-decoder agreement via exact identity and chemical/shape similarity in Figure~\ref{fig:cross_decoder_agreement}. The synthesis-decoded products preserve the 3D shape and pharmacophore characteristics implied by the 3D-decoded structures. We show examples of output pairs in Appendix~\ref{fig:app_syn_vs_3d_uncond}.



\subsection{Pharmacophore-conditioned molecule generation}

\paragraph{In-distribution pharmacophore conditioning.}

\begin{figure*}[h]
  \centering
  \begin{subfigure}[c]{0.5\textwidth}
      \centering
      \scriptsize 
      \setlength{\tabcolsep}{3pt}
      \renewcommand{\arraystretch}{1.1}

      \begin{tabular}{@{}lccccc@{}}
          \toprule
          \textbf{Method} &
          \multicolumn{1}{c}{\makecell[c]{\textbf{Hits}}} &
          \multicolumn{1}{c}{\makecell[c]{\textbf{Unique scaff.} \\ \textbf{hits}}} &
          \multicolumn{1}{c}{\makecell[c]{\textbf{Max} \\ \textbf{score}}} &
          \multicolumn{1}{c}{\makecell[c]{\textbf{AiZynth.} }} &
          \multicolumn{1}{c}{\makecell[c]{\textbf{Synthesizable} \\ \textbf{hits}}} \\
          \midrule
          Dataset baseline & 0 & 0 & 1.14 & 0.90 & 0 \\
          \rowcolor{Gray}
          \ours (3D out)   & 36.5 & 13.5 & 1.92 & 0.44 & 18.5 \\
          \rowcolor{Gray}
          \ours (syn out)  & 29 & 9.5 & 1.88 & 0.80 & 25.0 \\
          \bottomrule
      \end{tabular}
  \end{subfigure}%
  \hfill
  \begin{subfigure}[c]{0.5\textwidth}
      \centering
      \includegraphics[width=\linewidth]{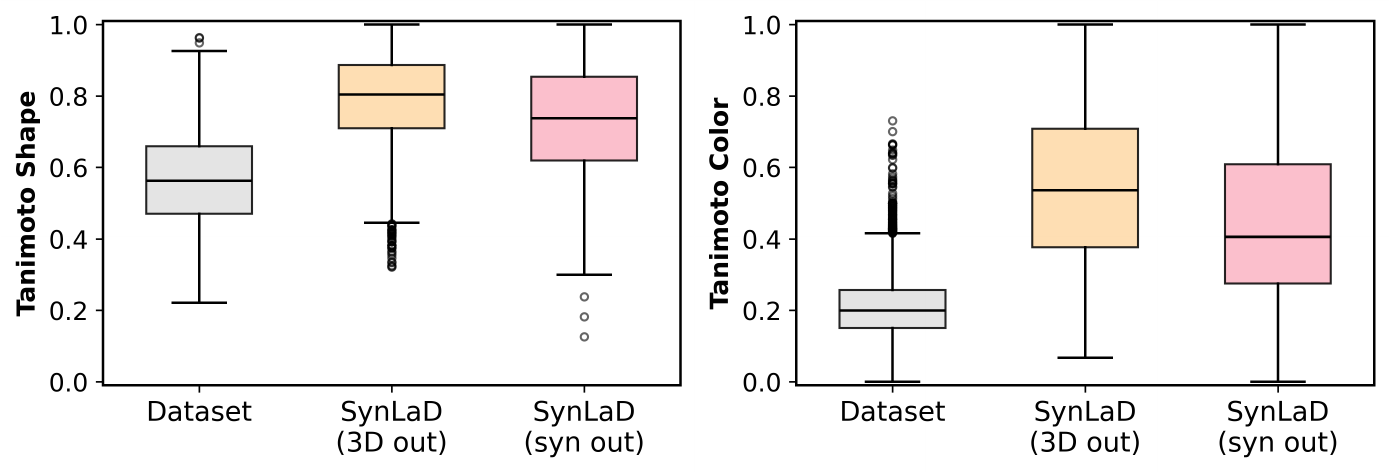}
  \end{subfigure}
  \vspace{-3pt}
  \caption{\textbf{Pharmacophore conditioned generation.} Metrics are evaluated for 50 random conditioning queries from a held-out set and 100 samples from \ours for each conditioning pharmacophore. \textit{Left}: Values for hits, unique scaffold hits, and max score are medians across the 50 query molecules, while AiZynth is a mean (higher is better). Max score represents the maximum Tanimoto combo score for generated samples. \textit{Right}: Distributions of ROCS shape and color (pharmacophore) similarity scores to query molecules (higher better).}
  \label{fig:pharm_cond}
  \vspace{-15pt}
\end{figure*}

We evaluate the conditional generation performance of \ours by selecting 50 random \textit{query} molecules from the test set and extracting their pharmacophores. For each query, we generate 100 candidate molecules with \ours conditioned on the extracted pharmacophore, and, as a baseline, compare against 5,000 randomly selected molecules from the training set. 
  All candidates---whether sampled from the training set or generated by \ours---are evaluated using conformer enumeration with Omega, followed by shape and pharmacophore overlay scoring against the query molecule (see Appendix~\ref{app:rocs_eval}). For consistency, we apply the same conformer enumeration to 3D-decoded outputs, even though these samples already exhibit good conformer alignment to the query (see Appendix Figure~\ref{fig:ph4_3d_examples}).
For each molecule, we retain the highest scoring conformer and assess whether it is a hit or not (i.e., with a Tanimoto combo score $\ge1.2$ to the query). We report the median number of hits and unique scaffold hits in Figure~\ref{fig:pharm_cond} and distributions of Tanimoto shape and color scores, as well as synthesizability scores assessed using AiZynthFinder (further metrics in Appendix Table~\ref{tab:app_id_pharmacophores}). Results show that \ours generates samples with significantly higher 3D pharmacophore and shape similarities to the query than the baseline, while the synthesis decoder yields molecules with high synthesizability.

\vspace{-8pt}

\paragraph{Screening case study.} To assess \ours in a practical drug-discovery setting, we compare it against brute-force ROCS screening over our dataset, which serves as a proxy for a molecular library and reflects standard but time-consuming ligand-based screening. We randomly select 10 query molecules with more than 15 heavy atoms from our test set and calculate ROCS scores against all other molecules in the dataset ($\sim$67k molecules), excluding the query molecules themselves. In parallel, we perform pharmacophore-conditional generation with \ours, generating \num{1000} samples per query. For this experiment, we consider only outputs of the synthesis decoder of \ours, ensuring that outputs are synthesizable and directly applicable for experimental testing. We report hit counts for both methods in Table~\ref{tab:dataset_screen}, and per-query results in Appendix Figure~\ref{app:fig_screen_vs_synlad}. \ours achieves higher hit counts than the typical brute-force baseline in ligand-based screening, while using nearly two orders of magnitude fewer samples (1k molecules) than the size of the screened library ($\sim$67k).

\begin{table}[h]
    \centering
    \footnotesize
    \caption{\textbf{Comparison to library screen.} Number of hits averaged over 10 queries from the test set. Here, we compare against the synthesis decoder of \ours as hits are more likely to be synthesizable and therefore able to be experimentally tested.}
    \label{tab:dataset_screen}
    \begin{tabular}{l c c}
        \hline
        \textbf{Method} & \textbf{Hits (avg.)} \\ 
        \hline
        Dataset screen & 49.7 \\ 
        \rowcolor{Gray}
        \ours & 59.4 \\ 
        \hline
    \end{tabular}
    \vspace{-10pt}
\end{table}

\begin{table*}[t]
    \centering
    \footnotesize
    \setlength{\tabcolsep}{3pt}
    \renewcommand{\arraystretch}{1.1}

    \caption{\textbf{Bioactive hit diversification.} We sample 500 samples per query for all methods. We report averaged metrics over all ten queries and AiZynthFinder success rate for a random subset of 100 generated molecules. Amortized methods do not require retraining for each target. Max score refers to the maximum Tanimoto combo score registered.}
    \label{tab:pharm_cond_results}

    \begin{tabular}{@{}lccccccc|c@{}}
        \toprule
        \textbf{Method} &
        \multicolumn{1}{c}{\makecell[c]{\textbf{Validity}}} &
        \multicolumn{1}{c}{\makecell[c]{\textbf{Hits (avg.)}}} &
        \multicolumn{1}{c}{\makecell[c]{\textbf{Unique scaff.} \\ \textbf{hits (avg.)}}} &
        \multicolumn{1}{c}{\makecell[c]{\textbf{Max} \\ \textbf{score}}} &
        \multicolumn{1}{c}{\makecell[c]{\textbf{AiZynth.}}} &
        \multicolumn{1}{c}{\makecell[c]{\textbf{Num. queries} \\
        \textbf{w/ $\boldsymbol{\geq1}$ hit}}} &
        \multicolumn{1}{c}{\makecell[c]{\textbf{Amortized}}} &
        \multicolumn{1}{|c}{\makecell[c]{\textbf{Synthesizable} \\ \textbf{hits (avg.)}}} \\
        \midrule
        REINVENT           & 0.98  & 21.1  & 3.7  & 1.15  & \textbf{0.75} & 3 & \xmark & \textbf{16.3} \\
        ShEPhERD           & 0.55  & 15.7 & 13.0 & 1.23 & 0.18 & 6 & \cmark & 6.3 \\
        SynFormer          & \textbf{1.0}  & 13.8 & 6.7  & 1.31  & 0.50 & \textbf{8} & \cmark &  6.7 \\ 
        \rowcolor{Gray}
        \rowcolor{Gray}
        \ours (3D out)     & 0.49  & \textbf{38.1} & \textbf{18.8} & \textbf{1.45} & 0.22 & \textbf{8} & \cmark & 12.7 \\
        \rowcolor{Gray}
        \ours (syn out)    & \textbf{1.0} & 17.9 & 6.9  & 1.30 & \textbf{0.75} & 5 & \cmark & 14.6 \\
        \bottomrule
    \end{tabular}
\end{table*}

\begin{figure*}[ht]
    \centering
    \includegraphics[width=0.95\textwidth]{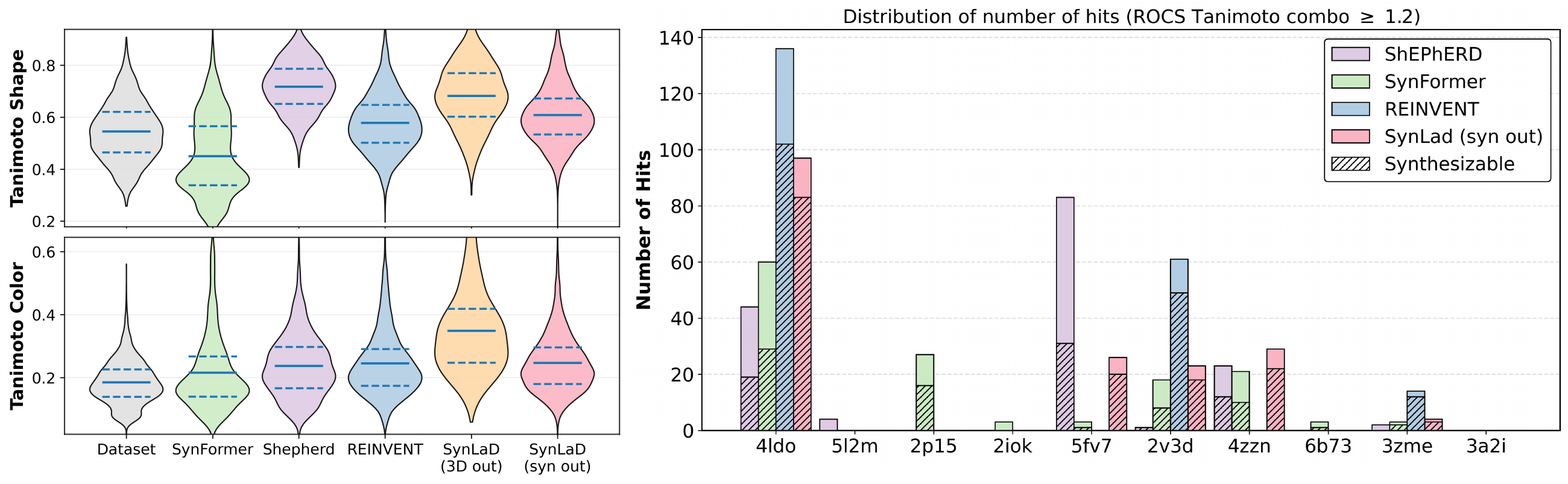}
    \caption{\textbf{Bioactive hit diversification.} Structural analogues are generated for 10 Lit-PCBA ligands using \ours and baseline methods. We report (1) the distribution of aggregated Tanimoto shape and color similarity scores to the query, and (2) the number of hits per query for each method, highlighting the fraction of synthesizable hits.}
    \label{fig:pharm_cond_litpcba}
\end{figure*}

\paragraph{Bioactive hit diversification.} We next investigate the out-of-distribution performance of \ours on a bioactive hit diversification task, which aims to find structural analogues of known bioactive compounds. This mirrors real-world drug discovery practices, where diverse candidates are needed to iteratively optimize activity while satisfying synthesizability, developability, and other constraints. We select ten targets from the Lit-PCBA benchmark \citep{litpcba} (see Appendix~\ref{app:pdb_exp}) and compare against ShEPhERD \citep{adamsShEPhERD2024}, SynFormer \citep{gao2022amortizedtreegenerationbottomup}, and REINVENT \citep{Blaschke2020,loeffler2024reinvent-f98}. ShEPhERD is an SE(3)-equivariant diffusion model that jointly denoises 3D molecular graphs and representations of their shapes and interaction profiles; we compare against their inpainting setting which conditions on electrostatic potential surfaces and pharmacophores. SynFormer is a transformer-based encoder-decoder framework that generates synthesizable 2D molecules in reaction-template and building-block space; we specifically compare against their SynFormerED variants for amortized analogue generation. As an orthogonal baseline, following \citet{PAPADOPOULOS2021116308}, we fine-tune REINVENT \citep{Blaschke2020} using a ROCS Tanimoto color score as reward function. Unlike \ours and the other baselines, REINVENT does not perform amortized sampling and instead requires reinforcement learning-based optimization per query, making the comparison methodologically orthogonal (similar techniques can also be applied to autoencoders \citep{tripp2020sample}). REINVENT is also substantially more computationally expensive, requiring approximately 6 hours per query to generate the optimized samples here, compared to $\sim$\SI{1}{min} for \ours, $\leq$\SI{1}{min} for SynFormer, and $\sim$\SI{28}{min} for ShEPhERD (Appendix~\ref{app:baselines}) per 100 samples, but we include it as it is a strong baseline in molecular generation.

For each query, we generate 500 molecules by conditioning on the pharmacophore profile of the native ligand in its bound conformation. We apply the same evaluation pipeline as in the previous section and report distributions of aggregated Tanimoto shape and color scores across all ten queries, along with averaged molecule validity, synthesizability, and hit counts, in Figure~\ref{fig:pharm_cond_litpcba} and Table~\ref{tab:pharm_cond_results}. Note that AiZynthFinder is reported for subsets (100 samples due to compute cost) of all generated molecules, not just hits. \ours achieves a strong overall balance, producing a high number of synthesizable hits, succeeding on a large fraction of queries, and maintaining high validity. Compared to the other synthesis-constrained baseline, SynFormer, \ours's synthesis decoder produces candidates with consistently higher ROCS shape and color overlap, suggesting that the explicit 3D representation our model learns improves conditional design; it also yields more hits and higher retrosynthesis success. The 3D outputs of \ours outperform ShEPhERD on most queries in terms of pharmacophore overlap, hit count, and synthesizability, while achieving comparable shape overlap. The synthesis-decoder output also outperforms ShEPhERD in average number of synthesizable hits.

Although REINVENT achieves the highest average number of synthesizable hits, this result is dominated by a single query with an unusually large hit count; across the benchmark, it generates synthesizable hits for only three queries and shows limited scaffold diversity. We show examples of query-generated molecule pairs with highlighted pharmacophore and shape overlap in Figure~\ref{fig:ph4_overlap} and further examples of synthesis and 3D decoder outputs in Appendix Figure~\ref{fig:app_examples_litpcba}. 

While we report results for both decoders for completeness, the key strength of our approach lies in the synthesis decoder, which---despite operating on linear synthesis trajectories---learns to satisfy shape and pharmacophore constraints more effectively than baselines that do not learn 3D information. This highlights the benefit of our strategy: learning latent representations that encode rich 3D conditioning signals and using them to condition synthesis generation. We further support this claim in Appendix~\ref{app:ablation}, where we perform two ablations: 1) we remove the structure-aware latent component (3D reconstruction heads and diffusion module) from \ours and keep the synthesis architecture and pharmacophore-conditioning fixed and 2) we switch the 3D reconstruction task for SMILES reconstruction, and keep the rest of the framework identical. We see that both ablations lead to 
a substantial performance drop, demonstrating that the structure-aware latent component is critical.

\begin{figure}[!h]
    \centering
    \includegraphics[width=\linewidth]{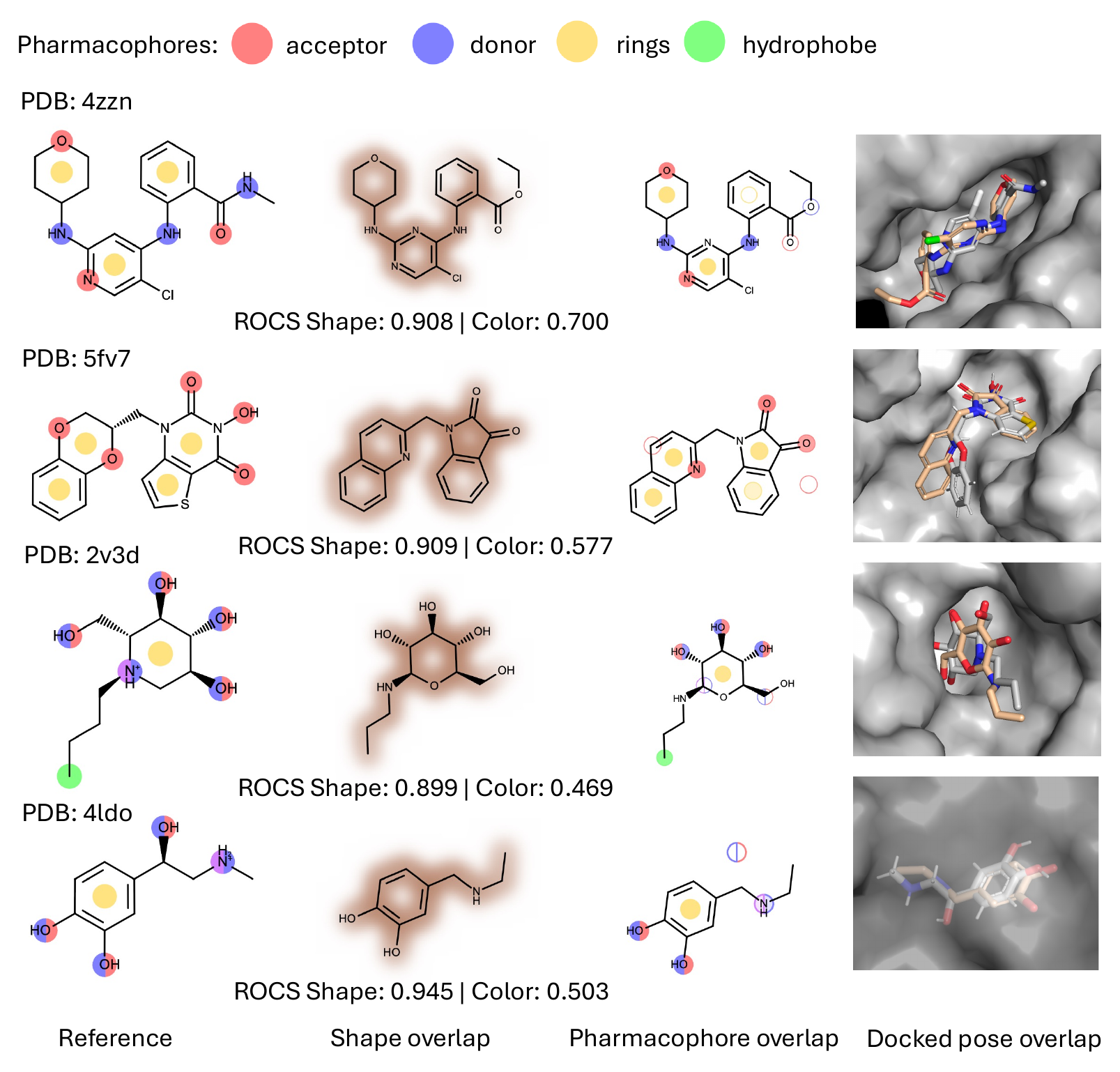}
    \caption{\textbf{Bioactive hit diversification experiment}. Examples of molecules generated by our model with annotated shape and pharmacophore overlap. The right-hand side shows the native ligand (gray) overlaid with the docked generated molecule (orange).}
    \label{fig:ph4_overlap}
\end{figure}

\section{Conclusion}

In this work, we introduce a conditional latent diffusion framework that decodes latent representations into both 3D molecular structures and reaction-based synthesis pathways. We show that jointly training these two decoders is mutually beneficial at sampling time: it increases the synthesizability of molecules produced by the 3D decoder and improves the synthesis decoder’s ability to satisfy shape/pharmacophore constraints under conditioning, while also enhancing sample diversity compared to a synthesis decoder conditioned directly on pharmacophores. Overall, our results demonstrate that conditioning an autoregressive synthesis decoder through a 3D-informed latent representation yields high numbers of diverse, synthesizable, pharmacophore-aligned hits. Looking ahead, an exciting direction is to scale our method to larger datasets (e.g., Pistachio, \citealp{NextMoveSoftware_Pistachio}), which would enable access to more novel transformations and more combinatorially complex regions of chemical space.

\section*{Impact Statement}

This paper presents a generative modeling approach aimed at advancing small-molecule drug discovery by jointly addressing 3D design objectives and synthetic feasibility. By enabling the generation of molecules that are both pharmacophore-consistent and synthesizable, our work has the potential to accelerate early-stage drug discovery and reduce reliance on costly and time-consuming experimental screening pipelines.

At the same time, we acknowledge that generative models for molecular design could, in principle, be misused to propose harmful or toxic compounds. While our work does not explicitly target such applications and is focused on drug discovery settings, the underlying methodology could be adapted in unintended ways.

We emphasize that our approach operates within standard drug discovery pipelines, where generated candidates are subject to extensive downstream validation, including safety, synthesizability, and biological evaluation. As such, we believe the primary impact of this work is to support responsible and beneficial applications in medicinal chemistry.

\FloatBarrier

\bibliography{icml2026_conference}

@inproceedings{rezende2015variational,
  title = 	 {Variational Inference with Normalizing Flows},
  author = 	 {Rezende, Danilo and Mohamed, Shakir},
  booktitle = 	 {Proceedings of the 32nd International Conference on Machine Learning},
  pages = 	 {1530--1538},
  year = 	 {2015},
  editor = 	 {Bach, Francis and Blei, David},
  volume = 	 {37},
  series = 	 {Proceedings of Machine Learning Research},
  address = 	 {Lille, France},
  month = 	 {07--09 Jul},
  publisher =    {PMLR},
  url = 	 {https://proceedings.mlr.press/v37/rezende15.html}
}

@inproceedings{lewis2020bart,
    title = "{BART}: Denoising Sequence-to-Sequence Pre-training for Natural Language Generation, Translation, and Comprehension",
    author = "Lewis, Mike  and
      Liu, Yinhan  and
      Goyal, Naman  and
      Ghazvininejad, Marjan  and
      Mohamed, Abdelrahman  and
      Levy, Omer  and
      Stoyanov, Veselin  and
      Zettlemoyer, Luke",
    editor = "Jurafsky, Dan  and
      Chai, Joyce  and
      Schluter, Natalie  and
      Tetreault, Joel",
    booktitle = "Proceedings of the 58th Annual Meeting of the Association for Computational Linguistics",
    month = jul,
    year = "2020",
    address = "Online",
    publisher = "Association for Computational Linguistics",
    url = "https://aclanthology.org/2020.acl-main.703/",
    pages = "7871--7880",
}

@inproceedings{wolf2019huggingface,
    title = "Transformers: State-of-the-Art Natural Language Processing",
    author = "Wolf, Thomas  and
      Debut, Lysandre  and
      Sanh, Victor  and
      Chaumond, Julien  and
      Delangue, Clement  and
      Moi, Anthony  and
      Cistac, Pierric  and
      Rault, Tim  and
      Louf, Remi  and
      Funtowicz, Morgan  and
      Davison, Joe  and
      Shleifer, Sam  and
      von Platen, Patrick  and
      Ma, Clara  and
      Jernite, Yacine  and
      Plu, Julien  and
      Xu, Canwen  and
      Le Scao, Teven  and
      Gugger, Sylvain  and
      Drame, Mariama  and
      Lhoest, Quentin  and
      Rush, Alexander",
    editor = "Liu, Qun  and
      Schlangen, David",
    booktitle = "Proceedings of the 2020 Conference on Empirical Methods in Natural Language Processing: System Demonstrations",
    month = oct,
    year = "2020",
    address = "Online",
    publisher = "Association for Computational Linguistics",
    url = "https://aclanthology.org/2020.emnlp-demos.6/",
    pages = "38--45"
}

@article{bradshaw2025challenging,
  title={Challenging reaction prediction models to generalize to novel chemistry},
  author={Bradshaw, John and Zhang, Anji and Mahjour, Babak and Graff, David E and Segler, Marwin HS and Coley, Connor W},
  journal={ACS Central Science},
  volume={11},
  number={4},
  pages={539--549},
  year={2025},
  publisher={ACS Publications},
  url={https://doi.org/10.1021/acscentsci.5c00055}
}

@inproceedings{maziarz2025chemist,
title={Chemist-aligned retrosynthesis by ensembling diverse inductive bias models},
author={Krzysztof Maziarz and Guoqing Liu and Austin Tripp and Junren Li and Piotr Gai{\'n}ski and Marwin Segler},
booktitle={NeurIPS 2025 AI for Science Workshop},
year={2025},
url={https://openreview.net/forum?id=F05pziBqOg}
}

@article{tu2022permutation,
  title={Permutation invariant graph-to-sequence model for template-free retrosynthesis and reaction prediction},
  author={Tu, Zhengkai and Coley, Connor W},
  journal={Journal of Chemical Information and Modeling},
  volume={62},
  number={15},
  pages={3503--3513},
  year={2022},
  publisher={ACS Publications},
  url={https://doi.org/10.1021/acs.jcim.2c00321}
}

@inproceedings{jin2017predicting,
 author = {Jin, Wengong and Coley, Connor and Barzilay, Regina and Jaakkola, Tommi},
 booktitle = {Advances in Neural Information Processing Systems},
 pages = {},
 editor = {I. Guyon and U. Von Luxburg and S. Bengio and H. Wallach and R. Fergus and S. Vishwanathan and R. Garnett},
 publisher = {Curran Associates, Inc.},
 title = {Predicting Organic Reaction Outcomes with {Weisfeiler-Lehman} Network},
 url = {https://proceedings.neurips.cc/paper_files/paper/2017/file/ced556cd9f9c0c8315cfbe0744a3baf0-Paper.pdf},
 volume = {30},
 year = {2017}
}

@inproceedings{
bradshaw2018generative,
title={A Generative Model For Electron Paths},
author={John Bradshaw and Matt J. Kusner and Brooks Paige and Marwin H. S. Segler and José Miguel Hernández-Lobato},
booktitle={International Conference on Learning Representations},
year={2019},
url={https://openreview.net/forum?id=r1x4BnCqKX},
}

@article{seidl2022improving,
  title={Improving few-and zero-shot reaction template prediction using modern hopfield networks},
  author={Seidl, Philipp and Renz, Philipp and Dyubankova, Natalia and Neves, Paulo and Verhoeven, Jonas and Wegner, Jorg K and Segler, Marwin and Hochreiter, Sepp and Klambauer, Gunter},
  journal={Journal of Chemical Information and Modeling},
  volume={62},
  number={9},
  pages={2111--2120},
  year={2022},
  publisher={ACS Publications},
  url={https://doi.org/10.1021/acs.jcim.1c01065}
}

@article{fooshee2018deep,
  title={Deep learning for chemical reaction prediction},
  author={Fooshee, David and Mood, Aaron and Gutman, Eugene and Tavakoli, Mohammadamin and Urban, Gregor and Liu, Frances and Huynh, Nancy and Van Vranken, David and Baldi, Pierre},
  journal={Molecular Systems Design \& Engineering},
  volume={3},
  number={3},
  pages={442--452},
  year={2018},
  publisher={Royal Society of Chemistry},
  url={https://doi.org/10.1039/C7ME00107J}
}

@article{joung2025electron,
  title={Electron flow matching for generative reaction mechanism prediction},
  author={Joung, Joonyoung F and Fong, Mun Hong and Casetti, Nicholas and Liles, Jordan P and Dassanayake, Ne S and Coley, Connor W},
  journal={Nature},
  volume={645},
  number={8079},
  pages={115--123},
  year={2025},
  publisher={Nature Publishing Group UK London},
  url={https://doi.org/10.1038/s41586-025-09426-9}
}

@article{kayala2011learning,
  title={Learning to predict chemical reactions},
  author={Kayala, Matthew A and Azencott, Chlo{\'e}-Agathe and Chen, Jonathan H and Baldi, Pierre},
  journal={Journal of Chemical Information and Modeling},
  volume={51},
  number={9},
  pages={2209--2222},
  year={2011},
  publisher={ACS Publications},
  url={https://doi.org/10.1021/ci200207y}
}

@article{sacha2021molecule,
  title={Molecule edit graph attention network: {Modeling} chemical reactions as sequences of graph edits},
  author={Sacha, Miko{\l}aj and B{\l}az, Miko{\l}aj and Byrski, Piotr and Dabrowski-Tumanski, Pawe{\l} and Chrominski, Miko{\l}aj and Loska, Rafa{\l} and W{\l}odarczyk-Pruszynski, Pawe{\l} and Jastrzebski, Stanis{\l}aw},
  journal={Journal of Chemical Information and Modeling},
  volume={61},
  number={7},
  pages={3273--3284},
  year={2021},
  publisher={ACS Publications},
  url={https://doi.org/10.1021/acs.jcim.1c00537}
}

@article{wei2016neural,
  title={Neural networks for the prediction of organic chemistry reactions},
  author={Wei, Jennifer N and Duvenaud, David and Aspuru-Guzik, Al{\'a}n},
  journal={ACS Central Science},
  volume={2},
  number={10},
  pages={725--732},
  year={2016},
  publisher={ACS Publications},
  url={https://doi.org/10.1021/acscentsci.6b00219}
}

@inproceedings{do2019graph,
author = {Do, Kien and Tran, Truyen and Venkatesh, Svetha},
title = {Graph Transformation Policy Network for Chemical Reaction Prediction},
year = {2019},
publisher = {Association for Computing Machinery},
address = {New York, NY, USA},
url = {https://doi.org/10.1145/3292500.3330958},
booktitle = {Proceedings of the 25th ACM SIGKDD International Conference on Knowledge Discovery \& Data Mining},
pages = {750--760},
numpages = {11},
location = {Anchorage, AK, USA},
series = {KDD '19}
}

@inproceedings{bi2021non,
  title = 	 {Non-Autoregressive Electron Redistribution Modeling for Reaction Prediction},
  author =       {Bi, Hangrui and Wang, Hengyi and Shi, Chence and Coley, Connor and Tang, Jian and Guo, Hongyu},
  booktitle = 	 {Proceedings of the 38th International Conference on Machine Learning},
  pages = 	 {904--913},
  year = 	 {2021},
  editor = 	 {Meila, Marina and Zhang, Tong},
  volume = 	 {139},
  series = 	 {Proceedings of Machine Learning Research},
  month = 	 {18--24 Jul},
  publisher =    {PMLR},
  url = 	 {https://proceedings.mlr.press/v139/bi21a.html}
}

@article{irwin2022chemformer,
  title={Chemformer: a pre-trained transformer for computational chemistry},
  author={Irwin, Ross and Dimitriadis, Spyridon and He, Jiazhen and Bjerrum, Esben Jannik},
  journal={Machine Learning: Science and Technology},
  volume={3},
  number={1},
  pages={015022},
  year={2022},
  publisher={IOP Publishing},
  url={https://doi.org/10.1088/2632-2153/ac3ffb}
}

@article{sagawa2025reactiont5,
  title={{ReactionT5}: a pre-trained transformer model for accurate chemical reaction prediction with limited data},
  author={Sagawa, Tatsuya and Kojima, Ryosuke},
  journal={Journal of Cheminformatics},
  volume={17},
  number={1},
  pages={126},
  year={2025},
  publisher={Springer},
  url={https://doi.org/10.1186/s13321-025-01075-4}
}

@misc{nam2016linking,
      title={Linking the Neural Machine Translation and the Prediction of Organic Chemistry Reactions}, 
      author={Juno Nam and Jurae Kim},
      year={2016},
      eprint={1612.09529},
      archivePrefix={arXiv},
      primaryClass={cs.LG},
      url={https://arxiv.org/abs/1612.09529}, 
}

@article{schwaller2018found,
  title={{“Found in Translation”}: predicting outcomes of complex organic chemistry reactions using neural sequence-to-sequence models},
  author={Schwaller, Philippe and Gaudin, Theophile and Lanyi, David and Bekas, Costas and Laino, Teodoro},
  journal={Chemical Science},
  volume={9},
  number={28},
  pages={6091--6098},
  year={2018},
  publisher={Royal Society of Chemistry},
  url={https://doi.org/10.1039/C8SC02339E}
}

@article{liu2017retrosynthetic,
  title={Retrosynthetic reaction prediction using neural sequence-to-sequence models},
  author={Liu, Bowen and Ramsundar, Bharath and Kawthekar, Prasad and Shi, Jade and Gomes, Joseph and Luu Nguyen, Quang and Ho, Stephen and Sloane, Jack and Wender, Paul and Pande, Vijay},
  journal={ACS Central Science},
  volume={3},
  number={10},
  pages={1103--1113},
  year={2017},
  publisher={ACS Publications},
  url={https://doi.org/10.1021/acscentsci.7b00303}
}

@article{bengio2003neural,
  title={A neural probabilistic language model},
  author={Bengio, Yoshua and Ducharme, R{\'e}jean and Vincent, Pascal and Jauvin, Christian},
  journal={Journal of Machine Learning Research},
  volume={3},
  pages={1137--1155},
  year={2003},
  url={https://dl.acm.org/doi/10.5555/944919.944966}
}

@inproceedings{sutskever2014sequence,
 author = {Sutskever, Ilya and Vinyals, Oriol and Le, Quoc V.},
 booktitle = {Advances in Neural Information Processing Systems},
 editor = {Z. Ghahramani and M. Welling and C. Cortes and N. Lawrence and K.Q. Weinberger},
 pages = {},
 publisher = {Curran Associates, Inc.},
 title = {Sequence to Sequence Learning with Neural Networks},
 url = {https://proceedings.neurips.cc/paper_files/paper/2014/file/5a18e133cbf9f257297f410bb7eca942-Paper.pdf},
 volume = {27},
 year = {2014}
}

@inproceedings{koziarski2024rgfn,
 author = {Koziarski, Micha\l  and Rekesh, Andrei and Shevchuk, Dmytro and van der Sloot, Almer and Gai\'{n}ski, Piotr and Bengio, Yoshua and Liu, Cheng-Hao and Tyers, Mike and Batey, Robert A.},
 booktitle = {Advances in Neural Information Processing Systems},
 editor = {A. Globerson and L. Mackey and D. Belgrave and A. Fan and U. Paquet and J. Tomczak and C. Zhang},
 pages = {46908--46955},
 publisher = {Curran Associates, Inc.},
 title = {{RGFN}: Synthesizable Molecular Generation Using {GFlowNets}},
 url = {https://doi.org/10.52202/079017-1488},
 volume = {37},
 year = {2024}
}

@inproceedings{korovina2020chembo,
  title = 	 {{ChemBO}: Bayesian Optimization of Small Organic Molecules with Synthesizable Recommendations},
  author =       {Korovina, Ksenia and Xu, Sailun and Kandasamy, Kirthevasan and Neiswanger, Willie and Poczos, Barnabas and Schneider, Jeff and Xing, Eric},
  booktitle = 	 {Proceedings of the Twenty Third International Conference on Artificial Intelligence and Statistics},
  pages = 	 {3393--3403},
  year = 	 {2020},
  editor = 	 {Chiappa, Silvia and Calandra, Roberto},
  volume = 	 {108},
  series = 	 {Proceedings of Machine Learning Research},
  month = 	 {26--28 Aug},
  publisher =    {PMLR},
  url = 	 {https://proceedings.mlr.press/v108/korovina20a.html}
}

@inproceedings{
lee2025rethinking,
title={Exploring Synthesizable Chemical Space with Iterative Pathway Refinements},
author={Seul Lee and Karsten Kreis and Srimukh Prasad Veccham and Meng Liu and Danny Reidenbach and Saee Gopal Paliwal and Weili Nie and Arash Vahdat},
booktitle={The Fourteenth International Conference on Learning Representations},
year={2026},
url={https://openreview.net/forum?id=aQKVfKOkR5}
}

@article{olivecrona2017molecular-7b7, 
  year     = {2017}, 
  title    = {Molecular de-novo design through deep reinforcement learning}, 
  author   = {Olivecrona, Marcus and Blaschke, Thomas and Engkvist, Ola and Chen, Hongming}, 
  journal  = {Journal of Cheminformatics}, 
  url      = {https://doi.org/10.1186/s13321-017-0235-x}, 
  pages    = {48}, 
  number   = {1}, 
  volume   = {9}
}

@article{mendez2019chembl-4bd, 
  year     = {2019}, 
  title    = {{ChEMBL}: towards direct deposition of bioassay data}, 
  author   = {Mendez, David and Gaulton, Anna and Bento, A Patrícia and Chambers, Jon and Veij, Marleen De and Félix, Eloy and Magariños, María Paula and Mosquera, Juan F and Mutowo, Prudence and Nowotka, Michał and Gordillo-Marañón, María and Hunter, Fiona and Junco, Laura and Mugumbate, Grace and Rodriguez-Lopez, Milagros and Atkinson, Francis and Bosc, Nicolas and Radoux, Chris J and Segura-Cabrera, Aldo and Hersey, Anne and Leach, Andrew R}, 
  journal  = {Nucleic Acids Research}, 
  url      = {https://doi.org/10.1093/nar/gky1075}, 
  pages    = {D930--D940}, 
  number   = {D1}, 
  volume   = {47}
}

@book{goodfellow2016deep,
    title={Deep Learning},
    author={Ian Goodfellow and Yoshua Bengio and Aaron Courville},
    publisher={MIT Press},
    url={http://www.deeplearningbook.org},
    year={2016}
}

@inproceedings{rombach2022highresolutionimagesynthesislatent,
author = { Rombach, Robin and Blattmann, Andreas and Lorenz, Dominik and Esser, Patrick and Ommer, Bjorn },
booktitle = { 2022 IEEE/CVF Conference on Computer Vision and Pattern Recognition (CVPR) },
title = {High-Resolution Image Synthesis with Latent Diffusion Models},
year = {2022},
volume = {},
pages = {10674--10685},
url = {https://doi.ieeecomputersociety.org/10.1109/CVPR52688.2022.01042},
publisher = {IEEE Computer Society},
address = {Los Alamitos, CA, USA},
month =Jun
}

@misc{kingma2022autoencodingvariationalbayes,
      title={Auto-Encoding Variational {Bayes}}, 
      author={Diederik P Kingma and Max Welling},
      year={2013},
      eprint={1312.6114},
      archivePrefix={arXiv},
      primaryClass={stat.ML},
      url={https://arxiv.org/abs/1312.6114}, 
}

@inproceedings{tripp2020sample,
 author = {Tripp, Austin and Daxberger, Erik and Hern\'{a}ndez-Lobato, Jos\'{e} Miguel},
 booktitle = {Advances in Neural Information Processing Systems},
 editor = {H. Larochelle and M. Ranzato and R. Hadsell and M.F. Balcan and H. Lin},
 pages = {11259--11272},
 publisher = {Curran Associates, Inc.},
 title = {Sample-Efficient Optimization in the Latent Space of Deep Generative Models via Weighted Retraining},
 url = {https://proceedings.neurips.cc/paper_files/paper/2020/file/81e3225c6ad49623167a4309eb4b2e75-Paper.pdf},
 volume = {33},
 year = {2020}
}

@inproceedings{Yu2024,
 author = {Yu, Kevin and Roh, Jihye and Li, Ziang and Gao, Wenhao and Wang, Runzhong and Coley, Connor W.},
 booktitle = {Advances in Neural Information Processing Systems},
 url = {https://doi.org/10.52202/079017-3588},
 editor = {A. Globerson and L. Mackey and D. Belgrave and A. Fan and U. Paquet and J. Tomczak and C. Zhang},
 pages = {112919--112949},
 publisher = {Curran Associates, Inc.},
 title = {Double-Ended Synthesis Planning with Goal-Constrained Bidirectional Search},
 volume = {37},
 year = {2024}
}

@article{segler2018generating,
  title={Generating focused molecule libraries for drug discovery with recurrent neural networks},
  author={Segler, Marwin HS and Kogej, Thierry and Tyrchan, Christian and Waller, Mark P},
  journal={ACS Central Science},
  volume={4},
  number={1},
  pages={120--131},
  year={2018},
  publisher={ACS Publications},
  url={https://doi.org/10.1021/acscentsci.7b00512},
}

@inproceedings{Peebles_2023_ICCV,
author = { Peebles, William and Xie, Saining },
booktitle = { 2023 IEEE/CVF International Conference on Computer Vision (ICCV) },
title = {Scalable Diffusion Models with Transformers},
year = {2023},
volume = {},
pages = {4172--4182},
url = {https://doi.ieeecomputersociety.org/10.1109/ICCV51070.2023.00387},
publisher = {IEEE Computer Society},
address = {Los Alamitos, CA, USA},
month =Oct
}

@inproceedings{joshi2025allatom,
  title={All-atom Diffusion Transformers: Unified generative modelling of molecules and materials},
  author={Chaitanya K. Joshi and Xiang Fu and Yi-Lun Liao and Vahe Gharakhanyan and Benjamin Kurt Miller and Anuroop Sriram and Zachary W. Ulissi},
  booktitle={International Conference on Machine Learning},
  year={2025},
  url={https://openreview.net/forum?id=89QPmZjIhv}
}

@inproceedings{
ho2022classifierfreediffusionguidance,
title={Classifier-Free Diffusion Guidance},
author={Jonathan Ho and Tim Salimans},
booktitle={NeurIPS 2021 Workshop on Deep Generative Models and Downstream Applications},
year={2021},
url={https://openreview.net/forum?id=qw8AKxfYbI}
}

@inproceedings{NIPS2017_3f5ee243,
 author = {Vaswani, Ashish and Shazeer, Noam and Parmar, Niki and Uszkoreit, Jakob and Jones, Llion and Gomez, Aidan N and Kaiser, Lukasz and Polosukhin, Illia},
 booktitle = {Advances in Neural Information Processing Systems},
 editor = {I. Guyon and U. Von Luxburg and S. Bengio and H. Wallach and R. Fergus and S. Vishwanathan and R. Garnett},
 pages = {},
 publisher = {Curran Associates, Inc.},
 title = {Attention is All you Need},
 url = {https://proceedings.neurips.cc/paper_files/paper/2017/file/3f5ee243547dee91fbd053c1c4a845aa-Paper.pdf},
 volume = {30},
 year = {2017}
}

@inproceedings{bradshaw_dog,
 author = {Bradshaw, John and Paige, Brooks and Kusner, Matt J and Segler, Marwin and Hern\'{a}ndez-Lobato, Jos\'{e} Miguel},
 booktitle = {Advances in Neural Information Processing Systems},
 editor = {H. Larochelle and M. Ranzato and R. Hadsell and M.F. Balcan and H. Lin},
 pages = {6852--6866},
 publisher = {Curran Associates, Inc.},
 title = {Barking up the right tree: {An} approach to search over molecule synthesis {DAGs}},
 url = {https://proceedings.neurips.cc/paper_files/paper/2020/file/4cc05b35c2f937c5bd9e7d41d3686fff-Paper.pdf},
 volume = {33},
 year = {2020}
}

@article{molecularTransformer,
author = {Schwaller, Philippe and Laino, Teodoro and Gaudin, Th{\'e}ophile and Bolgar, Peter and Hunter, Christopher A. and Bekas, Costas and Lee, Alpha A.},
title = {{Molecular Transformer}: A Model for Uncertainty-Calibrated Chemical Reaction Prediction},
journal = {ACS Central Science},
volume = {5},
number = {9},
pages = {1572--1583},
year = {2019},
URL = {https://doi.org/10.1021/acscentsci.9b00576}
}

@InProceedings{pmlr-v37-sohl-dickstein15,
  title = 	 {Deep Unsupervised Learning using Nonequilibrium Thermodynamics},
  author = 	 {Sohl-Dickstein, Jascha and Weiss, Eric and Maheswaranathan, Niru and Ganguli, Surya},
  booktitle = 	 {Proceedings of the 32nd International Conference on Machine Learning},
  pages = 	 {2256--2265},
  year = 	 {2015},
  editor = 	 {Bach, Francis and Blei, David},
  volume = 	 {37},
  series = 	 {Proceedings of Machine Learning Research},
  address = 	 {Lille, France},
  month = 	 {07--09 Jul},
  publisher =    {PMLR},
  url = 	 {https://proceedings.mlr.press/v37/sohl-dickstein15.html}
}

@inproceedings{Song2019GenerativeMB,
 author = {Song, Yang and Ermon, Stefano},
 booktitle = {Advances in Neural Information Processing Systems},
 editor = {H. Wallach and H. Larochelle and A. Beygelzimer and F. d\textquotesingle Alch\'{e}-Buc and E. Fox and R. Garnett},
 pages = {},
 publisher = {Curran Associates, Inc.},
 title = {Generative Modeling by Estimating Gradients of the Data Distribution},
 url = {https://proceedings.neurips.cc/paper_files/paper/2019/file/3001ef257407d5a371a96dcd947c7d93-Paper.pdf},
 volume = {32},
 year = {2019}
}

@inproceedings{Ho2020DenoisingDP,
 author = {Ho, Jonathan and Jain, Ajay and Abbeel, Pieter},
 booktitle = {Advances in Neural Information Processing Systems},
 editor = {H. Larochelle and M. Ranzato and R. Hadsell and M.F. Balcan and H. Lin},
 pages = {6840--6851},
 publisher = {Curran Associates, Inc.},
 title = {Denoising Diffusion Probabilistic Models},
 url = {https://proceedings.neurips.cc/paper_files/paper/2020/file/4c5bcfec8584af0d967f1ab10179ca4b-Paper.pdf},
 volume = {33},
 year = {2020}
}

@inproceedings{
Lipman2022FlowMF,
title={Flow Matching for Generative Modeling},
author={Yaron Lipman and Ricky T. Q. Chen and Heli Ben-Hamu and Maximilian Nickel and Matthew Le},
booktitle={The Eleventh International Conference on Learning Representations },
year={2023},
url={https://openreview.net/forum?id=PqvMRDCJT9t}
}

@misc{NextMoveSoftware_Pistachio,
  author       = {{NextMove Software}},
  title        = {Pistachio: Reaction Data, Querying and Analytics},
  year         = {2025},
  howpublished = {\url{https://www.nextmovesoftware.com/pistachio.html}},
  note         = {Accessed 2025-09-11}
}

@article{omega_quality,
author = {Friedrich, Nils-Ole and de Bruyn Kops, Christina and Flachsenberg, Florian and Sommer, Kai and Rarey, Matthias and Kirchmair, Johannes},
title = {Benchmarking Commercial Conformer Ensemble Generators},
journal = {Journal of Chemical Information and Modeling},
volume = {57},
number = {11},
pages = {2719--2728},
year = {2017},
URL = {https://doi.org/10.1021/acs.jcim.7b00505}
}

@inproceedings{jason_yim_selfcond,
author = {Yim, Jason and Trippe, Brian L. and De Bortoli, Valentin and Mathieu, Emile and Doucet, Arnaud and Barzilay, Regina and Jaakkola, Tommi},
title = {{SE(3)} diffusion model with application to protein backbone generation},
year = {2023},
publisher = {JMLR.org},
booktitle = {Proceedings of the 40th International Conference on Machine Learning},
articleno = {1672},
numpages = {39},
location = {Honolulu, Hawaii, USA},
series = {ICML'23},
url = {https://dl.acm.org/doi/10.5555/3618408.3620080}
}

@ARTICLE{moses,
  
AUTHOR={Polykovskiy, Daniil  and Zhebrak, Alexander  and Sanchez-Lengeling, Benjamin  and Golovanov, Sergey  and Tatanov, Oktai  and Belyaev, Stanislav  and Kurbanov, Rauf  and Artamonov, Aleksey  and Aladinskiy, Vladimir  and Veselov, Mark  and Kadurin, Artur  and Johansson, Simon  and Chen, Hongming  and Nikolenko, Sergey  and Aspuru-Guzik, Alán  and Zhavoronkov, Alex },
         
TITLE={{Molecular Sets (MOSES)}: A Benchmarking Platform for Molecular Generation Models},
        
JOURNAL={Frontiers in Pharmacology},
        
VOLUME={11},

YEAR={2020},

URL={https://doi.org/10.3389/fphar.2020.565644}
}

@article{buttenschoen2024posebusters,
  title = {{{PoseBusters}}: {{AI-based}} Docking Methods Fail to Generate Physically Valid Poses or Generalise to Novel Sequences},
  shorttitle = {{{PoseBusters}}},
  author = {Buttenschoen, Martin and Morris, Garrett M. and Deane, Charlotte M.},
  year = "2024",
  journal = "Chemical Science",
  volume = "15",
  issue = "9",
  pages = "3130-3139",
  publisher = "The Royal Society of Chemistry",
  url = "http://dx.doi.org/10.1039/D3SC04185A",
}

@article{sa_score,
title = {Estimation of synthetic accessibility score of drug-like molecules based on molecular complexity and fragment contributions},
volume = {1},
number = {8},
journal = {Journal of Cheminformatics},
author = {Ertl, Peter and Schuffenhauer, Ansgar},
year = {2009},
pages = {1--11},
url = {https://doi.org/10.1186/1758-2946-1-8}
}

@article{aizynth,
author = {Genheden, S. and Thakkar, A. and Chadimová, V. and Reymond, J. L. and Engkvist, O and Bjerrum, Esben},
title = {{AiZynthFinder}: a fast, robust and flexible open-source software for retrosynthetic planning},
journal = {Journal of Cheminformatics},
volume = {12},
pages = {70},
year = {2020},
url = {https://doi.org/10.1186/s13321-020-00472-1}
}

@article{Grebner2020,
  author  = {Grebner, Christoph and Malmerberg, Erik and Shewmaker, Andrew and Batista, Jose and Nicholls, Anthony and Sadowski, Jens},
  title   = {Virtual Screening in the Cloud: How Big Is Big Enough?},
  journal = {Journal of Chemical Information and Modeling},
  year    = {2020},
  volume  = {60},
  number  = {9},
  pages   = {4274--4282},
  url     = {https://doi.org/10.1021/acs.jcim.9b00779}
}

@inproceedings{adamsShEPhERD2024,
title={Sh{EP}h{ERD}: Diffusing shape, electrostatics, and pharmacophores for bioisosteric drug design},
author={Keir Adams and Kento Abeywardane and Jenna Fromer and Connor W. Coley},
booktitle={The Thirteenth International Conference on Learning Representations},
year={2025},
url={https://openreview.net/forum?id=KSLkFYHlYg}
}

@article{PAPADOPOULOS2021116308,
title = {De novo design with deep generative models based on {3D} similarity scoring},
journal = {Bioorganic \& Medicinal Chemistry},
volume = {44},
pages = {116308},
year = {2021},
url = {https://doi.org/10.1016/j.bmc.2021.116308},
author = {Kostas Papadopoulos and Kathryn A. Giblin and Jon Paul Janet and Atanas Patronov and Ola Engkvist}
}

@inproceedings{Higgins2016betaVAELB,
  title={beta-{VAE}: Learning Basic Visual Concepts with a Constrained Variational Framework},
  author={Irina Higgins and Lo{\"i}c Matthey and Arka Pal and Christopher P. Burgess and Xavier Glorot and Matthew M. Botvinick and Shakir Mohamed and Alexander Lerchner},
  booktitle={International Conference on Learning Representations},
  year={2017},
  url={https://openreview.net/forum?id=Sy2fzU9gl}
}

@article{Lowe2017,
author = "Daniel Lowe",
title = "{Chemical reactions from US patents (1976-Sep2016)}",
year = "2017",
month = "Jun",
url = "https://doi.org/10.6084/m9.figshare.5104873.v1"
}

@article{acharya2011recent,
  title={Recent advances in ligand-based drug design: Relevance and utility of the conformationally sampled pharmacophore approach},
  author={Acharya, C. and Coop, A. and Polli, J. E. and Mackerell, Jr., A. D.},
  journal={Current Computer-Aided Drug Design},
  volume={7},
  number={1},
  pages={10--22},
  year={2011},
  publisher={Bentham Science Publishers},
  url={https://doi.org/10.2174/157340911793743547}
}

@incollection{goodnow2007hit,
title = {1 {Hit} and Lead Identification: {Efficient} Practices for Drug Discovery},
editor = {F.D. King and G. Lawton},
series = {Progress in Medicinal Chemistry},
publisher = {Elsevier},
volume = {45},
pages = {1--61},
year = {2007},
url = {https://doi.org/10.1016/S0079-6468(06)45501-6},
author = {Goodnow, Jr., R. and Gillespie, P.}
}

@inproceedings{luo20223dgenerativemodelstructurebased,
 author = {Luo, Shitong and Guan, Jiaqi and Ma, Jianzhu and Peng, Jian},
 booktitle = {Advances in Neural Information Processing Systems},
 editor = {M. Ranzato and A. Beygelzimer and Y. Dauphin and P.S. Liang and J. Wortman Vaughan},
 pages = {6229--6239},
 publisher = {Curran Associates, Inc.},
 title = {A {3D} Generative Model for Structure-Based Drug Design},
 url = {https://proceedings.neurips.cc/paper_files/paper/2021/file/314450613369e0ee72d0da7f6fee773c-Paper.pdf},
 volume = {34},
 year = {2021}
}

@article{schneuing2024structurebaseddrugdesignequivariant,
  author  = {Schneuing, Arne and Harris, Charles and Du, Yuanqi and Didi, Kieran and Jamasb, Arian and Igashov, Ilia and Du, Weitao and Gomes, Carla and Blundell, Tom L. and Lio, Pietro and Welling, Max and Bronstein, Michael and Correia, Bruno},
  title   = {Structure-based drug design with equivariant diffusion models},
  journal = {Nature Computational Science},
  year    = {2024},
  volume  = {4},
  number  = {12},
  pages   = {899--909},
  url     = {https://doi.org/10.1038/s43588-024-00737-x}
}

@inproceedings{guan20233dequivariantdiffusiontargetaware,
title={{3D} Equivariant Diffusion for Target-Aware Molecule Generation and Affinity Prediction},
author={Jiaqi Guan and Wesley Wei Qian and Xingang Peng and Yufeng Su and Jian Peng and Jianzhu Ma},
booktitle={The Eleventh International Conference on Learning Representations},
year={2023},
url={https://openreview.net/forum?id=kJqXEPXMsE0}
}

@inproceedings{peng2025pocket2molefficientmolecularsampling,
  title = 	 {{P}ocket2{M}ol: Efficient Molecular Sampling Based on 3{D} Protein Pockets},
  author =       {Peng, Xingang and Luo, Shitong and Guan, Jiaqi and Xie, Qi and Peng, Jian and Ma, Jianzhu},
  booktitle = 	 {Proceedings of the 39th International Conference on Machine Learning},
  pages = 	 {17644--17655},
  year = 	 {2022},
  editor = 	 {Chaudhuri, Kamalika and Jegelka, Stefanie and Song, Le and Szepesvari, Csaba and Niu, Gang and Sabato, Sivan},
  volume = 	 {162},
  series = 	 {Proceedings of Machine Learning Research},
  month = 	 {17--23 Jul},
  publisher =    {PMLR},
  url = 	 {https://proceedings.mlr.press/v162/peng22b.html}
}

@Article{pilot,
author ="Cremer, Julian and Le, Tuan and Noé, Frank and Clevert, Djork-Arné and Schütt, Kristof T.",
title  ="{PILOT}: {Equivariant} diffusion for pocket-conditioned de novo ligand generation with multi-objective guidance via importance sampling",
journal  ="Chemical Science",
year  ="2024",
volume  ="15",
issue  ="36",
pages  ="14954--14967",
publisher  ="The Royal Society of Chemistry",
url  ="http://dx.doi.org/10.1039/D4SC03523B"}

@inproceedings{
schneuing2025multidomain,
title={Multi-domain Distribution Learning for De Novo Drug Design},
author={Arne Schneuing and Ilia Igashov and Adrian W. Dobbelstein and Thomas Castiglione and Michael M. Bronstein and Bruno Correia},
booktitle={The Thirteenth International Conference on Learning Representations},
year={2025},
url={https://openreview.net/forum?id=g3VCIM94ke}
}

@misc{cremer2025flowrflowmatchingstructureaware,
      title={{FLOWR}: Flow Matching for Structure-Aware De Novo, Interaction- and Fragment-Based Ligand Generation}, 
      author={Julian Cremer and Ross Irwin and Alessandro Tibo and Jon Paul Janet and Simon Olsson and Djork-Arné Clevert},
      year={2025},
      eprint={2504.10564},
      archivePrefix={arXiv},
      primaryClass={q-bio.QM},
      url={https://arxiv.org/abs/2504.10564}, 
}

@article{zhu2023pharmacophore,
  title={A pharmacophore-guided deep learning approach for bioactive molecular generation},
  author={Zhu, Huimin and Zhou, Renyi and Cao, Dongsheng and Tang, Jing and Li, Min},
  journal={Nature Communications},
  volume={14},
  number={1},
  pages={6234},
  year={2023},
  publisher={Nature Publishing Group UK},
  url={https://doi.org/10.1038/s41467-023-41454-9}
}

@article{STANLEY2023102658,
title = {Fake it until you make it? {Generative} de novo design and virtual screening of synthesizable molecules},
journal = {Current Opinion in Structural Biology},
volume = {82},
pages = {102658},
year = {2023},
url = {https://doi.org/10.1016/j.sbi.2023.102658},
author = {Megan Stanley and Marwin Segler}
}

@article{gao_synth,
author = {Gao, Wenhao and Coley, Connor W.},
title = {The Synthesizability of Molecules Proposed by Generative Models},
journal = {Journal of Chemical Information and Modeling},
volume = {60},
number = {12},
pages = {5714-5723},
year = {2020},
URL = {https://doi.org/10.1021/acs.jcim.0c00174}
}

@article{noutahi_synth,
author = {Horwood, Julien and Noutahi, Emmanuel},
title = {Molecular Design in Synthetically Accessible Chemical Space via Deep Reinforcement Learning},
journal = {ACS Omega},
volume = {5},
number = {51},
pages = {32984--32994},
year = {2020},
URL = {https://doi.org/10.1021/acsomega.0c04153},
}

@inproceedings{gao2022amortizedtreegenerationbottomup,
title={Amortized Tree Generation for Bottom-up Synthesis Planning and Synthesizable Molecular Design},
author={Wenhao Gao and Roc{\'\i}o Mercado and Connor W. Coley},
booktitle={International Conference on Learning Representations},
year={2022},
url={https://openreview.net/forum?id=FRxhHdnxt1}
}

@article{
gao2024generative,
author = {Wenhao Gao  and Shitong Luo  and Connor W. Coley },
title = {Generative {AI} for navigating synthesizable chemical space},
journal = {Proceedings of the National Academy of Sciences},
volume = {122},
number = {41},
pages = {e2415665122},
year = {2025},
URL = {https://www.pnas.org/doi/abs/10.1073/pnas.2415665122}
}

@misc{luo2025efficientprogrammableexplorationsynthesizable,
      title={Efficient and Programmable Exploration of Synthesizable Chemical Space}, 
      author={Shitong Luo and Connor W. Coley},
      year={2025},
      eprint={2512.00384},
      archivePrefix={arXiv},
      primaryClass={cs.LG},
      url={https://arxiv.org/abs/2512.00384}, 
}

@inproceedings{gottipati2020learning,
  title = 	 {Learning to Navigate The Synthetically Accessible Chemical Space Using Reinforcement Learning},
  author =       {Gottipati, Sai Krishna and Sattarov, Boris and Niu, Sufeng and Pathak, Yashaswi and Wei, Haoran and Liu, Shengchao and Liu, Shengchao and Blackburn, Simon and Thomas, Karam and Coley, Connor and Tang, Jian and Chandar, Sarath and Bengio, Yoshua},
  booktitle = 	 {Proceedings of the 37th International Conference on Machine Learning},
  pages = 	 {3668--3679},
  year = 	 {2020},
  editor = 	 {Daumé III, Hal and Singh, Aarti},
  volume = 	 {119},
  series = 	 {Proceedings of Machine Learning Research},
  month = 	 {13--18 Jul},
  publisher =    {PMLR},
  url = 	 {https://proceedings.mlr.press/v119/gottipati20a.html},
}

@article{swanson2025synthemol,
	author = {Swanson, Kyle and Liu, Gary and Catacutan, Denise B. and McLellan, Stewart and Arnold, Autumn and Tu, Megan M. and Brown, Eric D. and Zou, James and Stokes, Jonathan M.},
	title = {{SyntheMol-RL}: a flexible reinforcement learning framework for designing novel and synthesizable antibiotics},
	year = {2025},
	url = {https://doi.org/10.1101/2025.05.17.654017},
	publisher = {Cold Spring Harbor Laboratory},
	journal = {bioRxiv}
}

@article{gobbi,
	author = {Gobbi, Alberto and Poppinger, Dieter},
	title = {Genetic optimization of combinatorial libraries.},
	journal = {Biotechnology and Bioengineering},
    year = {1998},
    volume = {61},
    number = {1},
    pages = {47--54},
    doi = {10.1002/(sici)1097-0290(199824)61:1<47::aid-bit9>3.0.co;2-z},
}

@misc{lo2025genetic,
  title={A Genetic Algorithm for Navigating Synthesizable Molecular Spaces},
  author={Lo, Alston and Coley, Connor W and Matusik, Wojciech},
  eprint={2509.20719},
  archivePrefix={arXiv},
  primaryClass={cs.LG},
  url={https://arxiv.org/abs/2509.20719}, 
  year={2025}
}

@article{hartenfeller2012dogs-a10, 
  year     = {2012}, 
  title    = {{DOGS}: Reaction-Driven de novo Design of Bioactive Compounds}, 
  author   = {Hartenfeller, Markus and Zettl, Heiko and Walter, Miriam and Rupp, Matthias and Reisen, Felix and Proschak, Ewgenij and Weggen, Sascha and Stark, Holger and Schneider, Gisbert}, 
  journal  = {{PLoS} Computational Biology}, 
  url      = {https://doi.org/10.1371/journal.pcbi.1002380}, 
  pages    = {e1002380}, 
  number   = {2}, 
  volume   = {8}
}

@article{Vinkers2003, 
  year     = {2003}, 
  title    = {{SYNOPSIS}: {SYNthesize} and {OPtimize} System in Silico}, 
  author   = {Vinkers, H. Maarten and Jonge, Marc R. de and Daeyaert, Frederik F. D. and Heeres, Jan and Koymans, Lucien M. H. and Lenthe, Joop H. van and Lewi, Paul J. and Timmerman, Henk and Aken, Koen Van and Janssen, Paul A. J.}, 
  journal  = {Journal of Medicinal Chemistry}, 
  url      = {https://doi.org/10.1021/jm030809x}, 
  pages    = {2765--2773}, 
  number   = {13}, 
  volume   = {46}
}

@article{liu2022retrognn-714, 
  year     = {2022}, 
  title    = {{RetroGNN}: Fast Estimation of Synthesizability for Virtual Screening and De Novo Design by Learning from Slow Retrosynthesis Software}, 
  author   = {Liu, Cheng-Hao and Korablyov, Maksym and Jastrzebski, Stanisław and Włodarczyk-Pruszy\'nski, Paweł and Bengio, Yoshua and Segler, Marwin}, 
  journal  = {Journal of Chemical Information and Modeling}, 
  url      = {https://doi.org/10.1021/acs.jcim.1c01476}, 
  pmid     = {35452226},
  pages    = {2293--2300}, 
  number   = {10}, 
  volume   = {62}
}

@article{guo2025directly-aea, 
  year     = {2025}, 
  title    = {Directly optimizing for synthesizability in generative molecular design using retrosynthesis models}, 
  author   = {Guo, Jeff and Schwaller, Philippe}, 
  journal  = {Chemical Science}, 
  url      = {https://doi.org/10.1039/d5sc01476j}, 
  pages    = {6943--6956}, 
  issue   = {16}, 
  volume   = {16}
}

@article{gomezbombarelli2018automatic-f5e, 
  year     = {2018}, 
  title    = {Automatic Chemical Design Using a Data-Driven Continuous Representation of Molecules}, 
  author   = {Gómez-Bombarelli, Rafael and Wei, Jennifer N. and Duvenaud, David and Hernández-Lobato, José Miguel and Sánchez-Lengeling, Benjamín and Sheberla, Dennis and Aguilera-Iparraguirre, Jorge and Hirzel, Timothy D. and Adams, Ryan P. and Aspuru-Guzik, Alán}, 
  journal  = {{ACS} Central Science}, 
  url      = {https://doi.org/10.1021/acscentsci.7b00572}, 
  pages    = {268--276}, 
  number   = {2}, 
  volume   = {4}
}

@article{swanson2024generative,
  title={Generative {AI} for designing and validating easily synthesizable and structurally novel antibiotics},
  author={Swanson, Kyle and Liu, Gary and Catacutan, Denise B. and Arnold, Autumn and Zou, James and Stokes, Jonathan M.},
  journal={Nature Machine Intelligence},
  volume={6},
  number={3},
  pages={338--353},
  year={2024},
  publisher={Nature Publishing Group},
  url={https://doi.org/10.1038/s42256-024-00809-7}
}

@inproceedings{NEURIPS2019_46d0671d,
 author = {Bradshaw, John and Paige, Brooks and Kusner, Matt J and Segler, Marwin and Hern\'{a}ndez-Lobato, Jos\'{e} Miguel},
 booktitle = {Advances in Neural Information Processing Systems},
 editor = {H. Wallach and H. Larochelle and A. Beygelzimer and F. d\textquotesingle Alch\'{e}-Buc and E. Fox and R. Garnett},
 pages = {},
 publisher = {Curran Associates, Inc.},
 title = {A Model to Search for Synthesizable Molecules},
 url = {https://proceedings.neurips.cc/paper_files/paper/2019/file/46d0671dd4117ea366031f87f3aa0093-Paper.pdf},
 volume = {32},
 year = {2019}
}

@inproceedings{
cretu2025synflownetdesigndiversenovel,
title={{SynFlowNet}: {Design} of Diverse and Novel Molecules with Synthesis Constraints},
author={Miruna Cretu and Charles Harris and Ilia Igashov and Arne Schneuing and Marwin Segler and Bruno Correia and Julien Roy and Emmanuel Bengio and Pietro Lio},
booktitle={The Thirteenth International Conference on Learning Representations},
year={2025},
url={https://openreview.net/forum?id=uvHmnahyp1}
}

@InProceedings{luo2025projectingmoleculessynthesizablechemical,
  title = 	 {Projecting Molecules into Synthesizable Chemical Spaces},
  author =       {Luo, Shitong and Gao, Wenhao and Wu, Zuofan and Peng, Jian and Coley, Connor W. and Ma, Jianzhu},
  booktitle = 	 {Proceedings of the 41st International Conference on Machine Learning},
  pages = 	 {33289--33304},
  year = 	 {2024},
  editor = 	 {Salakhutdinov, Ruslan and Kolter, Zico and Heller, Katherine and Weller, Adrian and Oliver, Nuria and Scarlett, Jonathan and Berkenkamp, Felix},
  volume = 	 {235},
  series = 	 {Proceedings of Machine Learning Research},
  month = 	 {21--27 Jul},
  publisher =    {PMLR},
  url = 	 {https://proceedings.mlr.press/v235/luo24a.html}
}

@InProceedings{liu2023audioldmtexttoaudiogenerationlatent,
  title = 	 {{A}udio{LDM}: Text-to-Audio Generation with Latent Diffusion Models},
  author =       {Liu, Haohe and Chen, Zehua and Yuan, Yi and Mei, Xinhao and Liu, Xubo and Mandic, Danilo and Wang, Wenwu and Plumbley, Mark D},
  booktitle = 	 {Proceedings of the 40th International Conference on Machine Learning},
  pages = 	 {21450--21474},
  year = 	 {2023},
  editor = 	 {Krause, Andreas and Brunskill, Emma and Cho, Kyunghyun and Engelhardt, Barbara and Sabato, Sivan and Scarlett, Jonathan},
  volume = 	 {202},
  series = 	 {Proceedings of Machine Learning Research},
  month = 	 {23--29 Jul},
  publisher =    {PMLR},
  url = 	 {https://proceedings.mlr.press/v202/liu23f.html},
}

@article{litpcba,
author = {Tran-Nguyen, Viet-Khoa and Jacquemard, C{\'e}lien and Rognan, Didier},
title = {{LIT-PCBA}: An Unbiased Data Set for Machine Learning and Virtual Screening},
journal = {Journal of Chemical Information and Modeling},
volume = {60},
number = {9},
pages = {4263--4273},
year = {2020},
URL = {https://doi.org/10.1021/acs.jcim.0c00155}
}

@article{omega,
author = {Hawkins, Paul C. D. and Skillman, A. Geoffrey and Warren, Gregory L. and Ellingson, Benjamin A. and Stahl, Matthew T.},
title = {Conformer Generation with {OMEGA}: Algorithm and Validation Using High Quality Structures from the Protein Databank and Cambridge Structural Database},
journal = {Journal of Chemical Information and Modeling},
volume = {50},
number = {4},
pages = {572--584},
year = {2010},
URL = {https://doi.org/10.1021/ci100031x},
}

@article{bemis_murcko,
author = {Bemis, Guy W. and Murcko, Mark A.},
title = {The Properties of Known Drugs. 1. {M}olecular Frameworks},
journal = {Journal of Medicinal Chemistry},
volume = {39},
number = {15},
pages = {2887--2893},
year = {1996},
URL = {https://doi.org/10.1021/jm9602928}
}

@article{chem_space_growth,
author = {Kuan, Jacqueline and Radaeva, Mariia and Avenido, Adeline and Cherkasov, Artem and Gentile, Francesco},
title = {Keeping pace with the explosive growth of chemical libraries with structure-based virtual screening},
journal = {WIREs Computational Molecular Science},
volume = {13},
number = {6},
pages = {e1678},
url = {https://doi.org/10.1002/wcms.1678},
year = {2023}
}

@misc{rocs_web,
  author       = {{OpenEye Scientific Software}},
  title        = {{ROCS}: Shape Similarity},
  year         = {2025},
  howpublished = {\url{https://www.eyesopen.com/rocs}},
  note         = {Accessed 2025-01-20}
}

@article{rocs_eg,
author = {Hawkins, Paul C. D. and Skillman, A. Geoffrey and Nicholls, Anthony},
title = {Comparison of Shape-Matching and Docking as Virtual Screening Tools},
journal = {Journal of Medicinal Chemistry},
volume = {50},
number = {1},
pages = {74--82},
year = {2007},
URL = {https://doi.org/10.1021/jm0603365}
}

@article{rocs_eg2,
  author  = {Sheridan, Robert P. and McGaughey, Georgia B. and Cornell, Wendy D.},
  title   = {Multiple protein structures and multiple ligands: effects on the apparent goodness of virtual screening results},
  journal = {Journal of Computer-Aided Molecular Design},
  year    = {2008},
  volume  = {22},
  number  = {3},
  pages   = {257--265},
  url     = {https://doi.org/10.1007/s10822-008-9168-9},
}

@Article{D1SC02436A,
author ="Imrie, Fergus and Hadfield, Thomas E. and Bradley, Anthony R. and Deane, Charlotte M.",
title  ="Deep generative design with {3D} pharmacophoric constraints",
journal  ="Chemical Science",
year  ="2021",
volume  ="12",
issue  ="43",
pages  ="14577--14589",
publisher  ="The Royal Society of Chemistry",
url  ="http://dx.doi.org/10.1039/D1SC02436A",
}

@article{xie2024acceleratingdiscoverynovelbioactive,
  author  = {Xie, Weixin and Zhang, Jianhang and Xie, Qin and Gong, Chaojun and Ren, Yuhao and Xie, Jin and Sun, Qi and Xu, Youjun and Lai, Luhua and Pei, Jianfeng},
  title   = {Accelerating discovery of bioactive ligands with pharmacophore-informed generative models},
  journal = {Nature Communications},
  year    = {2025},
  volume  = {16},
  number  = {1},
  pages   = {2391},
  url     = {https://doi.org/10.1038/s41467-025-56349-0},
}

@misc{mahmood2025pharmacophorebaseddesignlearningvoxel,
      title={Pharmacophore-based design by learning on voxel grids}, 
      author={Omar Mahmood and Pedro O. Pinheiro and Richard Bonneau and Saeed Saremi and Vishnu Sresht},
      year={2025},
      eprint={2512.02031},
      archivePrefix={arXiv},
      primaryClass={cs.LG},
      url={https://arxiv.org/abs/2512.02031}, 
}

@article{ziv_molsnapper,
author = {Ziv, Yael and Imrie, Fergus and Marsden, Brian and Deane, Charlotte M.},
title = {{MolSnapper}: Conditioning Diffusion for Structure-Based Drug Design},
journal = {Journal of Chemical Information and Modeling},
volume = {65},
number = {9},
pages = {4263--4273},
year = {2025},
URL = {https://doi.org/10.1021/acs.jcim.4c02008}
}

@inproceedings{pinheiro20243dmoleculegenerationdenoising,
 author = {O. Pinheiro, Pedro O and Rackers, Joshua and Kleinhenz, Joseph and Maser, Michael and Mahmood, Omar and Watkins, Andrew and Ra, Stephen and Sresht, Vishnu and Saremi, Saeed},
 booktitle = {Advances in Neural Information Processing Systems},
 editor = {A. Oh and T. Naumann and A. Globerson and K. Saenko and M. Hardt and S. Levine},
 pages = {69077--69097},
 publisher = {Curran Associates, Inc.},
 title = {{3D} molecule generation by denoising voxel grids},
 url = {https://proceedings.neurips.cc/paper_files/paper/2023/file/da1131a86ac3c70e0b7cae89c3d4df22-Paper-Conference.pdf},
 volume = {36},
 year = {2023}
}

@inproceedings{
shen2025compositionalflows3dmolecule,
title={Compositional Flows for {3D} Molecule and Synthesis Pathway Co-design},
author={Tony Shen and Seonghwan Seo and Ross Irwin and Kieran Didi and Simon Olsson and Woo Youn Kim and Martin Ester},
booktitle={Forty-second International Conference on Machine Learning},
year={2025},
url={https://openreview.net/forum?id=4aXfSLfM0Z}
}

@misc{rekesh2025syncogensynthesizable3dmolecule,
      title={{SynCoGen}: Synthesizable {3D} Molecule Generation via Joint Reaction and Coordinate Modeling}, 
      author={Andrei Rekesh and Miruna Cretu and Dmytro Shevchuk and Vignesh Ram Somnath and Pietro Liò and Robert A. Batey and Mike Tyers and Michał Koziarski and Cheng-Hao Liu},
      year={2025},
      eprint={2507.11818},
      archivePrefix={arXiv},
      primaryClass={cs.LG},
      url={https://arxiv.org/abs/2507.11818}, 
}

@inproceedings{NEURIPS2021_5dca4c6b,
 author = {Vahdat, Arash and Kreis, Karsten and Kautz, Jan},
 booktitle = {Advances in Neural Information Processing Systems},
 editor = {M. Ranzato and A. Beygelzimer and Y. Dauphin and P.S. Liang and J. Wortman Vaughan},
 pages = {11287--11302},
 publisher = {Curran Associates, Inc.},
 title = {Score-based Generative Modeling in Latent Space},
 url = {https://proceedings.neurips.cc/paper_files/paper/2021/file/5dca4c6b9e244d24a30b4c45601d9720-Paper.pdf},
 volume = {34},
 year = {2021}
}

@misc{xu2023geometriclatentdiffusionmodels,
      title={Geometric Latent Diffusion Models for {3D} Molecule Generation}, 
      author={Minkai Xu and Alexander Powers and Ron Dror and Stefano Ermon and Jure Leskovec},
      year={2023},
      eprint={2305.01140},
      archivePrefix={arXiv},
      primaryClass={cs.LG},
      url={https://arxiv.org/abs/2305.01140}, 
}

@article{Gupta2018GenerativeRNN,
  author  = {Gupta, A. and M{\"u}ller, A. T. and Huisman, B. J. H. and Fuchs, J. A. and Schneider, P. and Schneider, G.},
  title   = {Generative Recurrent Networks for De Novo Drug Design},
  journal = {Molecular Informatics},
  year    = {2018},
  volume  = {37},
  number  = {1--2},
  pages   = {1700111},
  url     = {https://doi.org/10.1002/minf.201700111}
}

@article{Blaschke2020,
author = {Blaschke, Thomas and Arús-Pous, Josep and Chen, Hongming and Margreitter, Christian and Tyrchan, Christian and Engkvist, Ola and Papadopoulos, Kostas and Patronov, Atanas},
title = {{REINVENT 2.0}: An {AI} Tool for De Novo Drug Design},
journal = {Journal of Chemical Information and Modeling},
volume = {60},
number = {12},
pages = {5918--5922},
year = {2020},
URL = {https://doi.org/10.1021/acs.jcim.0c00915}
}

@article{loeffler2024reinvent-f98, 
  year     = {2024}, 
  title    = {Reinvent 4: Modern {AI}–driven generative molecule design}, 
  author   = {Loeffler, Hannes H. and He, Jiazhen and Tibo, Alessandro and Janet, Jon Paul and Voronov, Alexey and Mervin, Lewis H. and Engkvist, Ola}, 
  journal  = {Journal of Cheminformatics}, 
  url      = {https://doi.org/10.1186/s13321-024-00812-5},
  pages    = {20}, 
  number   = {1}, 
  volume   = {16}
}

@Article{C8SC05372C,
author ="Jensen, Jan H.",
title  ="A graph-based genetic algorithm and generative model/{Monte Carlo} tree search for the exploration of chemical space",
journal  ="Chemical Science",
year  ="2019",
volume  ="10",
issue  ="12",
pages  ="3567--3572",
publisher  ="The Royal Society of Chemistry",
url  ="http://dx.doi.org/10.1039/C8SC05372C"
}

@inproceedings{
qin2025defogdiscreteflowmatching,
title={{DeFoG}: Discrete Flow Matching for Graph Generation},
author={Yiming QIN and Manuel Madeira and Dorina Thanou and Pascal Frossard},
booktitle={Forty-second International Conference on Machine Learning},
year={2025},
url={https://openreview.net/forum?id=KPRIwWhqAZ}
}

@misc{vonessen2025tabascofastsimplifiedmodel,
      title={TABASCO: A Fast, Simplified Model for Molecular Generation with Improved Physical Quality}, 
      author={Carlos Vonessen and Charles Harris and Miruna Cretu and Pietro Liò},
      year={2025},
      eprint={2507.00899},
      archivePrefix={arXiv},
      primaryClass={cs.LG},
      url={https://arxiv.org/abs/2507.00899}, 
}

@inproceedings{irwin2025semlaflowefficient3d,
  title={{SemlaFlow} -- {Efficient} {3D} Molecular Generation with Latent Attention and Equivariant Flow Matching},
  author={Ross Irwin and Alessandro Tibo and Jon Paul Janet and Simon Olsson},
  booktitle={The 28th International Conference on Artificial Intelligence and Statistics},
  year={2025},
  url={https://openreview.net/forum?id=bee2G6pEh0}
}

@inproceedings{
le2023navigatingdesignspaceequivariant,
title={Navigating the Design Space of Equivariant Diffusion-Based Generative Models for De Novo {3D} Molecule Generation},
author={Tuan Le and Julian Cremer and Frank Noe and Djork-Arn{\'e} Clevert and Kristof T Sch{\"u}tt},
booktitle={The Twelfth International Conference on Learning Representations},
year={2024},
url={https://openreview.net/forum?id=kzGuiRXZrQ}
}

@misc{dunn2024mixedcontinuouscategoricalflow,
      title={Mixed Continuous and Categorical Flow Matching for {3D} De Novo Molecule Generation}, 
      author={Ian Dunn and David Ryan Koes},
      year={2024},
      eprint={2404.19739},
      archivePrefix={arXiv},
      primaryClass={q-bio.BM},
      url={https://arxiv.org/abs/2404.19739}, 
}

@inproceedings{huang2025midimultiinstancediffusionsingle,
  title={{MIDI}: Multi-instance diffusion for single image to {3D} scene generation},
  author={Huang, Zehuan and Guo, Yuan-Chen and An, Xingqiao and Yang, Yunhan and Li, Yangguang and Zou, Zi-Xin and Liang, Ding and Liu, Xihui and Cao, Yan-Pei and Sheng, Lu},
  booktitle={Proceedings of the Computer Vision and Pattern Recognition Conference},
  pages={23646--23657},
  month={June},
  year={2025},
  url={https://openaccess.thecvf.com/content/CVPR2025/papers/Huang_MIDI_Multi-Instance_Diffusion_for_Single_Image_to_3D_Scene_Generation_CVPR_2025_paper.pdf}
}

@article{ONG2013314,
title = {{Python Materials Genomics} (pymatgen): A robust, open-source python library for materials analysis},
journal = {Computational Materials Science},
volume = {68},
pages = {314--319},
year = {2013},
url = {https://doi.org/10.1016/j.commatsci.2012.10.028},
author = {Shyue Ping Ong and William Davidson Richards and Anubhav Jain and Geoffroy Hautier and Michael Kocher and Shreyas Cholia and Dan Gunter and Vincent L. Chevrier and Kristin A. Persson and Gerbrand Ceder}
}

@inproceedings{
seo2025generativeflowssyntheticpathway,
title={Generative Flows on Synthetic Pathway for Drug Design},
author={Seonghwan Seo and Minsu Kim and Tony Shen and Martin Ester and Jinkyoo Park and Sungsoo Ahn and Woo Youn Kim},
booktitle={The Thirteenth International Conference on Learning Representations},
year={2025},
url={https://openreview.net/forum?id=pB1XSj2y4X}
}
\bibliographystyle{icml2026}

\newpage
\appendix
\onecolumn

\section{Limitations and Future Work}

The focus of this work is to introduce and validate \ours's ability to generate small molecules that preserve key phramacophore interactions of ligands and have high synthesizability. To this end, we restrict our methodology and experiments to ligand-based pharmacophore conditioning, reaction-constrained synthesis-pathway generation, and evaluation on small-molecule analogue generation tasks, as well as evaluation of our method's core modules. A current limitation of our framework is that it relies on external reaction predicitons, and its performance is therefore affected by predictor bias or failure models. Future work could evaluate alternative predictors or ensembles of predictors. Second, our experiments here are limited to the USPTO dataset, and scaling to larger and more diverse reaction datasets may further test the generality of the approach. Additionally, further work could also focus on improving cross-decoder agreement (e.g., perhaps via considering techniques from contrastive learning), as stronger consistency across decoding heads may lead to better molecule feature alignment.

\section{Methods}

\subsection{Dataset}
\label{app:dataset}

For our dataset, we make use of the provided synthesis DAGs from \citet{bradshaw_dog}. 
Briefly, these DAGs were obtained from a cleaned version of the USPTO reaction dataset after stripping out reagents \citep{jin2017predicting, Lowe2017}; first, by building up a reaction network from the USPTO reactions and an initial set of ``building block'' nodes (picked by frequency), before extracting a possible synthesis plan for each non-building block node by tracing reactions backward to a loop-free subgraph that terminates in building blocks (selecting a single route when multiple alternatives exist).
Using this data, we first filter the reactions such that the final products contain between 10 and 40 heavy atoms, and obtain \num{67512} synthesis DAGs, 90\% of which we keep as our training set and use the remaining data for our validation and test sets.
Note that we randomly split by final molecule (i.e., the different train, validation, and test sets lead to different final molecules but may share reactions/sub-networks); analyzing \ours's ability to extrapolate to different data splits (for instance, a ``Reachable'' vs ``Hard'' split akin to \citealp[p.7]{Yu2024}) is an interesting future direction to explore.
For each final product of the synthesis DAGs we compute a maximum of 10 low-energy conformers using Omega \citep{omega}, which results in a total of \num{584896} 3D molecules. 

\subsection{Reaction network serialization}
\label{app:networkSerialization}

\begin{figure}[ht]
    \centering
    \includegraphics[width=\linewidth]{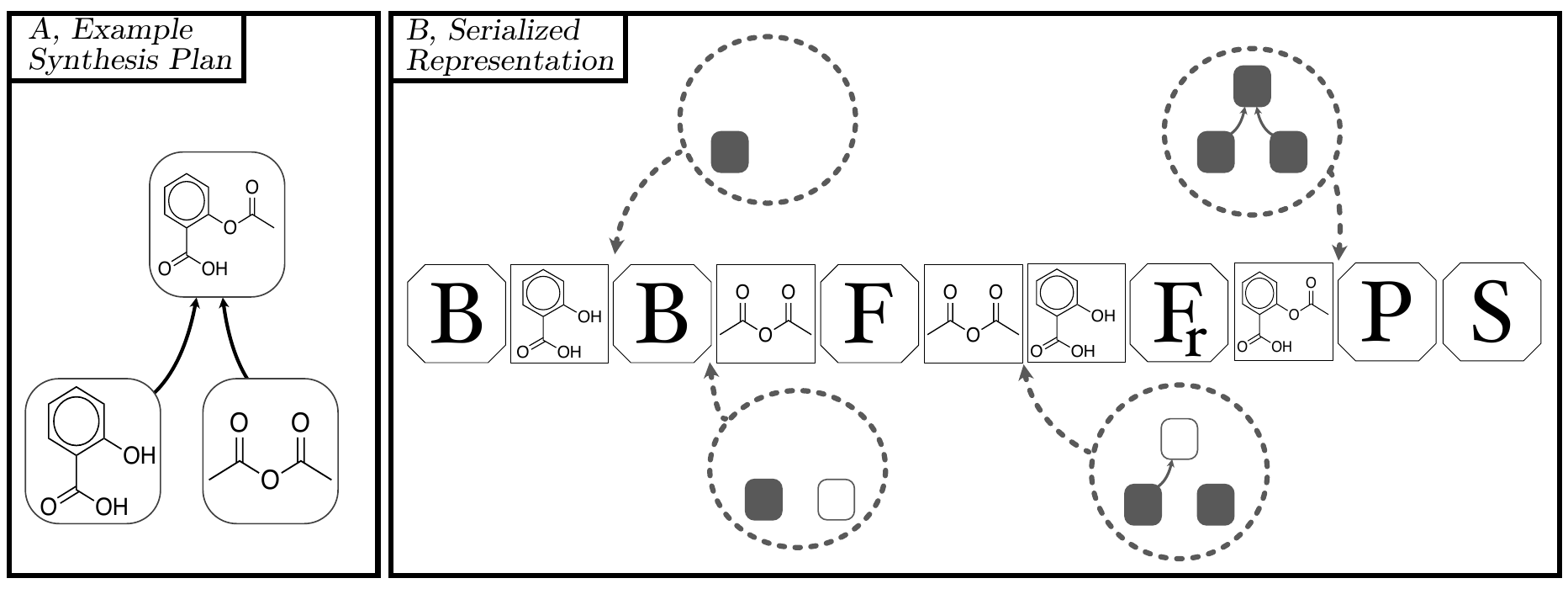}
    \caption{A tokenization scheme for describing synthesis plans in a bottom-up manner similar to \citet{bradshaw_dog}. The synthesis plan, depicted as a DAG in \textbf{A} can be serialized as shown in \textbf{B} into a series of tokens.
    In \textbf{B}, the four dotted circles above the sequence indicate the state of the synthesis plan at that stage of decoding---empty nodes indicate the identity of the molecule corresponding to that node has yet to be defined. Note that this scheme can be used to describe complex, multi-step synthesis plans in a bottom-up manner.
    }
    \label{fig:networkSerialization}
\end{figure}

A synthesis pathway here is defined as a directed acyclic graph (DAG), where nodes represent molecules (each unique molecule maps to a unique node) and the edges represent reactions. In order to describe these pathways using an autoregressive language model we need to serialize them into a linear sequence of tokens. Our proposed approach  follows \citep{bradshaw_dog} (alternative serialization schemes have also been proposed, \citealp{gao2024generative,lee2025rethinking,luo2025projectingmoleculessynthesizablechemical}) and is shown in Figure~\ref{fig:networkSerialization}.
The scheme relies on two kinds of tokens: \textit{action tokens} and \textit{molecular tokens}. Action tokens define a new operation to take with respect to the graph and are generally followed by a series of molecular nodes that specify the arguments required to define the action's effect. There are three types of action tokens:

\begin{itemize}
\item  \texttt{`B'} adds a building block node to the network (these are easily purchasable compounds and the token is followed by a single molecule token defining the identity of the chosen building block).

\item  \texttt{`F'} adds a reaction edge (and a product node if the molecule does not yet exist in the network) to our network through a single forward reaction. This action is followed by selecting one or more molecule nodes (representing the existing molecules in the network) to use as reactants, a sub-action \texttt{`F_r'} (which indicates the reaction should be run) and a final product node (followed by \texttt{`P'}) that represents the product of the reaction. 

\item \texttt{`S'} is the stop action, which halts the prediction.  
\end{itemize}

\subsection{Synthesis action embeddings} 
\label{app:actionEmbeddings}

The input tokens to the transformer are obtained via summing a series of embeddings as shown in Figure~\ref{fig:modelOverview}C: a learned embedding which specifies the token type, a molecule embedding (if the action involves a molecule) which is a projection of the molecule's 2048-bit Morgan fingerprint, and a learned positional embedding. 

\subsection{VAE training}
\label{app:vae_training}

\paragraph{Autoencoder ablations} We report ablations for autoencoder hyperparameters in Tables~\ref{tab:vae_weights} and~\ref{tab:app_autoencoder_reconstruct}. Increasing the weight on the synthesis decoder loss (see Eq.~\eqref{eq:loss}) harms 3D coordinate reconstruction and adds little benefit for the synthesis sequence reconstruction. Note that we report the accuracy of builiding block prediction in its order-invariant form (i.e., building blocks might be added earlier or later on in the sequence, with the end product being the same); this differs during training in which we randomly pick an order (for building blocks at the same level) and train on this fixed sequence using teacher forcing. We use $\lambda=0.92$ in our experiments and $\beta=\num{e-5}$, as larger KL weight harms reconstruction accuracy for both decoders. We use a latent dimension $d=16$ which shows the best molecule and synthesis reconstruction.

\begin{table}[]
    \centering
    \footnotesize
    \caption{\textbf{Autoencoder loss weights ablation study.} We report 3D molecule match rate (computed with MoleculeMatcher from PyMatGen) and invariant building block match rate in the synthesis sequence.}
    \begin{tabular}{c c c c}
    \toprule
          & &
        {3D molecule} & {Building block} \\
        {$\lambda$} &  $\beta$ & {match rate} & {match rate} \\
        \midrule
        $0.50$ & $\num{e-5}$ & 82.1 & 83.3 \\
        $0.70$ & $\num{e-5}$ & 90.2 & 83.1 \\
        $0.92$ & $\num{e-5}$ & 98.5 & 82.5 \\
        $0.98$ & $\num{e-5}$ & 98.7 & 80.6 \\
        $0.92$ & $\num{e-4}$ & 80.6 & 67.4 \\
        \bottomrule
    \end{tabular}
    \label{tab:vae_weights}
\end{table}

\begin{table*}
  \centering
  \footnotesize
  \setlength{\tabcolsep}{2.8pt}
  \renewcommand{\arraystretch}{1.15}

  \caption{\textbf{Autoencoder reconstruction accuracies.} The synthesis match rate and Morgan fingerprint Tanimoto similarity are calculated between the final product of the predicted and the ground truth synthesis sequences. We report performance with two different inference strategies for the synthesis decoder.}
  \label{tab:app_autoencoder_reconstruct}
  
  \begin{tabularx}{\textwidth}{*{2}{Y}*2{Y}*2{Y}} 
    \toprule
    \multicolumn{2}{c}{\textbf{Autoencoder hyperparameters}} & 
    \multicolumn{2}{c}{\textbf{3D metrics}} & 
    \multicolumn{2}{c}{\textbf{Synthesis metrics}} \\
    \cmidrule(r){1-2} \cmidrule(r){3-4} \cmidrule{5-6}
    
    {Inference method} & 
    {Latent dim.} & 
    {Match Rate (\%)}$\uparrow$ & 
    {RMSD (\AA) $\downarrow$} & 
    {Match Rate (\%)}$\uparrow$ & 
    {Tanimoto similarity}$\uparrow$ \\
    \midrule
    Beam search & 8 & 98.9 & 0.05 & 58.4 & 0.82 \\
    Beam search & 16 & 98.5 & 0.05 & 63.4 & 0.84 \\
    Beam search & 32 & 98.9 & 0.04 & 62.7 & 0.83 \\
    \midrule
    Sampling & 8 & 98.9 & 0.05 & 42.8 & 0.73 \\
    Sampling & 16 & 98.5 & 0.05 & 49.0 & 0.76 \\
    Sampling & 32 & 98.5 & 0.04 & 51.4 & 0.77 \\
    \bottomrule
  \end{tabularx}
\end{table*}

\begin{table*}[t]
\centering
\footnotesize
\setlength{\tabcolsep}{1.5pt}        
\renewcommand{\arraystretch}{1.0}    

\begin{minipage}[t]{0.49\textwidth}
\centering
\caption{{\textbf{Synthesis decoder ablations (beam search).}}}
\label{tab:app_beam_search_ablations}
\begin{tabularx}{\linewidth}{YYY}
\toprule
{Beam width} &
{Synthesis Match Rate (\%)$\uparrow$} &
{Tanimoto similarity$\uparrow$} \\
\midrule
{1}  & {47.6} & {0.75} \\
{5}  & {63.4} & {0.84} \\
{10} & {59.5} & {0.82} \\
\bottomrule
\end{tabularx}
\end{minipage}\hfill
\begin{minipage}[t]{0.49\textwidth}
\centering
\caption{{\textbf{Synthesis decoder ablations (sampling).}}}
\label{tab:app_sampling_ablations}
\begin{tabularx}{\linewidth}{ccYY}
\toprule
{N} &
{Top-$k$} &
{Synthesis Match Rate (\%)$\uparrow$} &
{Tanimoto similarity$\uparrow$} \\
\midrule
{1}  & {1}  & {43.9} & {0.73} \\
{1}  & {5}  & {44.1} & {0.73} \\
{1}  & {10} & {43.4} & {0.73} \\
\bottomrule
\end{tabularx}
\end{minipage}
\end{table*}

\subsubsection{Synthesis decoder training} 
\label{app:synthesisTraining}

The synthesis decoder is implemented as a decoder-only transformer. Each input token is embedded by summing (i) a learned embedding for special tokens or retrieved 
molecular embeddings, (ii) a learned token-type embedding, and (iii) a learned positional embedding. Graph-level information is provided via a fingerprint encoder. The embedded sequence is processed by transformer blocks with RMSNorm pre-normalization, multi-head self-attention, and a SwiGLU feed-forward network (further details can be found in our attached code). When conditioning is enabled, each block additionally applies a cross-attention sublayer over an external conditioning vector. During training, we use teacher forcing \citep[\S10.2.1]{goodfellow2016deep}, i.e., for each prediction step the decoder is conditioned on the ground-truth previous tokens. We compute the cross-entropy loss only on positions where the model is expected to generate tokens, and mask out product tokens (and any padded positions) so they do not contribute to the loss. We report ablations for synthesis decoder inference in Tables~\ref{tab:app_beam_search_ablations} and~\ref{tab:app_sampling_ablations}.

\subsection{DiT training}
\label{app:dit_training}

\paragraph{Sampling ablations.} We investigate the effect of the number of sampling steps used during inference on molecule quality and synthesis/3D decoder output match rate in Table~\ref{tab:num_steps_ablation}. Unless otherwise stated, we use 100 sampling steps in all experiments. We additionally ablate the classifier-free guidance weight $w$ in Eq.~\eqref{eq:cfg} (Table~\ref{tab:cfg_ablation}). Increasing $w$ improves control over latent quality and yields more valid decoded molecules, but reduces scaffold diversity. Since our primary objective is to maximize the number of generated hits, we therefore use $w=0$ (i.e., no classifier-free guidance) throughout.

\begin{table}
  \centering
  \footnotesize
  \setlength{\tabcolsep}{6pt}
  \caption{\textbf{Number of sampling steps ablation.} We performed the ablation in the unconditional setting.}
  \renewcommand{\arraystretch}{1.15}
  \begin{tabular}{rcccc}
    \toprule
    \#Steps & Validity$\uparrow$ & PoseBusters$\uparrow$ & Synthesis/3D exact match rate$\uparrow$ \\
    \midrule
     50  & 0.85 & 0.85 & 0.28 \\
    100  & 0.89 & 0.89 & 0.33 \\
    200  & 0.89 & 0.89 & 0.32 \\
    300  & 0.90 & 0.89& 0.33 \\
    500  & 0.91 & 0.89 & 0.33 \\
    \bottomrule
  \end{tabular}
  \label{tab:num_steps_ablation}
\end{table}

\begin{table}
  \centering
  \footnotesize
  \setlength{\tabcolsep}{6pt}
  \caption{\textbf{Classifier free guidance ablation.} We performed the ablation for the Lit-PCBA ligands conditioning experiment and generated 100 samples per conditioning input.}
  \renewcommand{\arraystretch}{1.15}
  \begin{tabular}{r c cc cc cc}
    \toprule
    cfg. setting &
    Validity$\uparrow$ &
    \multicolumn{2}{c}{Diversity$\uparrow$} &
    \multicolumn{2}{c}{Hits$\uparrow$} &
    \multicolumn{2}{c}{Uniq. scaff. hits$\uparrow$} \\
    \cmidrule(lr){3-4}\cmidrule(lr){5-6}\cmidrule(lr){7-8}
     & &
    (syn out) & (3D out) &
    (syn out) & (3D out) &
    (syn out) & (3D out) \\
    \midrule
     w/o cfg.  & 0.49 & 0.86 & 0.86 & 17.9 & 38.1 & 6.9 & 18.8 \\
     $w=0.5$   & 0.56 & 0.85 & 0.85 & 14.2 & 34.7 & 3.8 & 12.1 \\
     $w=1.0$   & 0.46 & 0.85 & 0.84 & 14.3 & 38.0 & 4.8 & 15.2 \\
     $w=2.0$   & 0.45 & 0.85 & 0.84 & 13.1 & 34.7 & 5.0 & 15.7 \\
     $w=4.0$   & 0.36 & 0.85 & 0.84 & 11.7 & 23.4 & 5.2 & 12.6 \\
    \bottomrule
  \end{tabular}
  \label{tab:cfg_ablation}
\end{table}

\subsection{Training and hyperparameters}
\label{app:hyperparam}

We jointly train the VAE and synthesis decoder, then train the DiT in a second stage, using AdamW with a constant learning rate of $\num{e-4}$, zero weight decay, and a batch size of 256. We maintain an exponential moving average of the DiT parameters with decay $0.9999$. Each stage is run to convergence, for up to 3000 epochs or 3 days on a single B200 GPU.

For the VAE, we use standard transformers for both the encoder and 3D decoder with hidden dimension $d_{\text{model}}=256$, 4 attention heads and 6 layers. For the synthesis decoder we used  $d_{\text{model}}=384$, 8 attention heads, and 8 layers with causal self-attention, cross-attention conditioning to the latents, and SwigLU FFNs. 

For the second stage training, the pharmacophore encoder is a standard transformer with $d_{\text{model}}=256$, 4 heads, and 4 layers, and the DiT denoiser has $d_{\text{model}}=768$, 12 attention heads, and 12 layers.

\paragraph{Molecule and pharmacophore embeddings.} Molecules and pharmacophores are represented by their respective atom or pharmacophore types together with 3D coordinates (see Sections \ref{section:encoder} and \ref{section:conditioning_method}). We use separate embedding layers for atom / pharmacophore types and atomic positions, following \citet{joshi2025allatom}. Atom types are embedded via a learned embedding lookup, while positions are projected using a linear layer. The resulting representations are combined by summation.

\subsection{Reaction predictor oracle training} 
\label{app:reactionOracle}

As explained in the main text, when running our synthesis decoder at inference time we require access to a reaction predictor oracle to predict the molecular identity of the product molecule given the set of reactants (we ignore conditions and reagents in this current work). 
For this, we use a model based around the BART architecture \citep{lewis2020bart}, using an implementation adapted from the \texttt{transformers} library \citep{wolf2019huggingface} for use in reaction prediction \citep{bradshaw2025challenging}. 
This is similar in spirit to other encoder-decoder models used for reaction prediction/one-step retrosynthesis \citep{nam2016linking, molecularTransformer,schwaller2018found,liu2017retrosynthetic}.
We train our reaction prediction oracle on reactant-product pairs from the same USPTO dataset we extract the synthesis plans from for training \ours; teacher forcing is used at training time and beam search is used during inference for making predictions. 
We leave to future work the exploration of using alternative reaction predictors \citep{tu2022permutation,jin2017predicting,bradshaw2018generative,seidl2022improving,joung2025electron,fooshee2018deep,kayala2011learning, sacha2021molecule,wei2016neural,do2019graph,bi2021non,irwin2022chemformer,sagawa2025reactiont5}, or even multiple at a time \citep{maziarz2025chemist} for more robust predictions. We report accuracy results from training on a train/validation/test split of approximately 409k/30k/40k data points extracted from the USPTO dataset in Table~\ref{tab:app_rxn_predictor}.

\begin{table}[t]
\centering
\footnotesize
\setlength{\tabcolsep}{4pt}
\renewcommand{\arraystretch}{1.15}
\caption{\textbf{Reaction predictor test accuracy.}
Top-k accuracy (\%) on the held-out test set for product outcome prediction.}
\label{tab:app_rxn_predictor}
\begin{tabular}{lcccc}
\toprule
\textbf{Checkpoint} & \textbf{Top-1} & \textbf{Top-2} & \textbf{Top-3} & \textbf{Top-5} \\
\midrule
Final (1M iters) & 84.6 & 91.4 & 93.3 & 94.8 \\
\bottomrule
\end{tabular}
\end{table}

\subsection{Evaluation metrics}
\label{app:metrics}

We define the metrics used thoroughout the paper as follows:
\begin{itemize}
    \item \textbf{Validity (Valid.)}: \% of molecules that can be processed by RDKit.
    \item \textbf{Internal diversity (IntDiv.)}: One minus the value of the averaged pairwise Tanimoto similarities of the Morgan fingerprints (size 2048, radius 2) of the molecules in the generated set.
    \item \textbf{Murcko Scaffold count}: number of unique Murcko scaffolds in a generated set.
    \item \textbf{Novelty (Nov.)}: \% of valid molecules not present in the training set.
    \item \textbf{PoseBusters (PB)}: \% of molecules passing all PoseBusters filters \citep{buttenschoen2024posebusters}. This was only evaluated for models that generated 3D poses.
    \item \textbf{SA}: Averaged synthetic accessibility score from \citet{sa_score}.
    \item \textbf{AiZynthFinder (AiZynth.)}: \% of molecules (out of 100) for which AiZynthFinder finds valid synthetic pathways given its default set of reactions and ZINC building blocks.  
    \item \textbf{RMSD and match rate}: For 3D molecule reconstruction we report RMSD and match rate computed using MoleculeMatcher from PyMatGen \citep{ONG2013314}.
    \item \textbf{Hit rate}: We compute hit rate by counting the number of unique molecules with ROCS Tanimoto Combo score $\geq 1.2$.
\end{itemize}

\subsection{ROCS evaluation pipeline}
\label{app:rocs_eval}

We scored 3D similarity with OpenEye ROCS \citep{rocs_eg} using a standard OMEGA→ROCS workflow. We kept the query conformation intact (from our test set or Lit-PCBA benchmark pose) and for each candidate molecule, we first generated an ensemble of 100 low-energy 3D conformations with OMEGA \citep{omega}, which constructs initial geometries from fragment conformers and then samples torsions using a knowledge-based torsion library, retaining a pruned set of energetically and geometrically reasonable conformers. We then used ROCS to perform fast rigid overlays between all query-candidate conformer pairs, representing molecular shape as a sum of atom-centered Gaussian functions and maximizing shared volume and pharmacophoric (``color'') feature overlap. For each candidate, we reported the best overlay across conformers, and computed ShapeTanimoto, ColorTanimoto, and their sum (TanimotoCombo) as the final shape, color, and combined similarity scores. The same pipeline was applied for \ours and all baselines, regardless of the methods' output types.

\FloatBarrier
\section{Experiments}

\begin{figure}[p]
    \centering
    \includegraphics[width=0.9\linewidth]{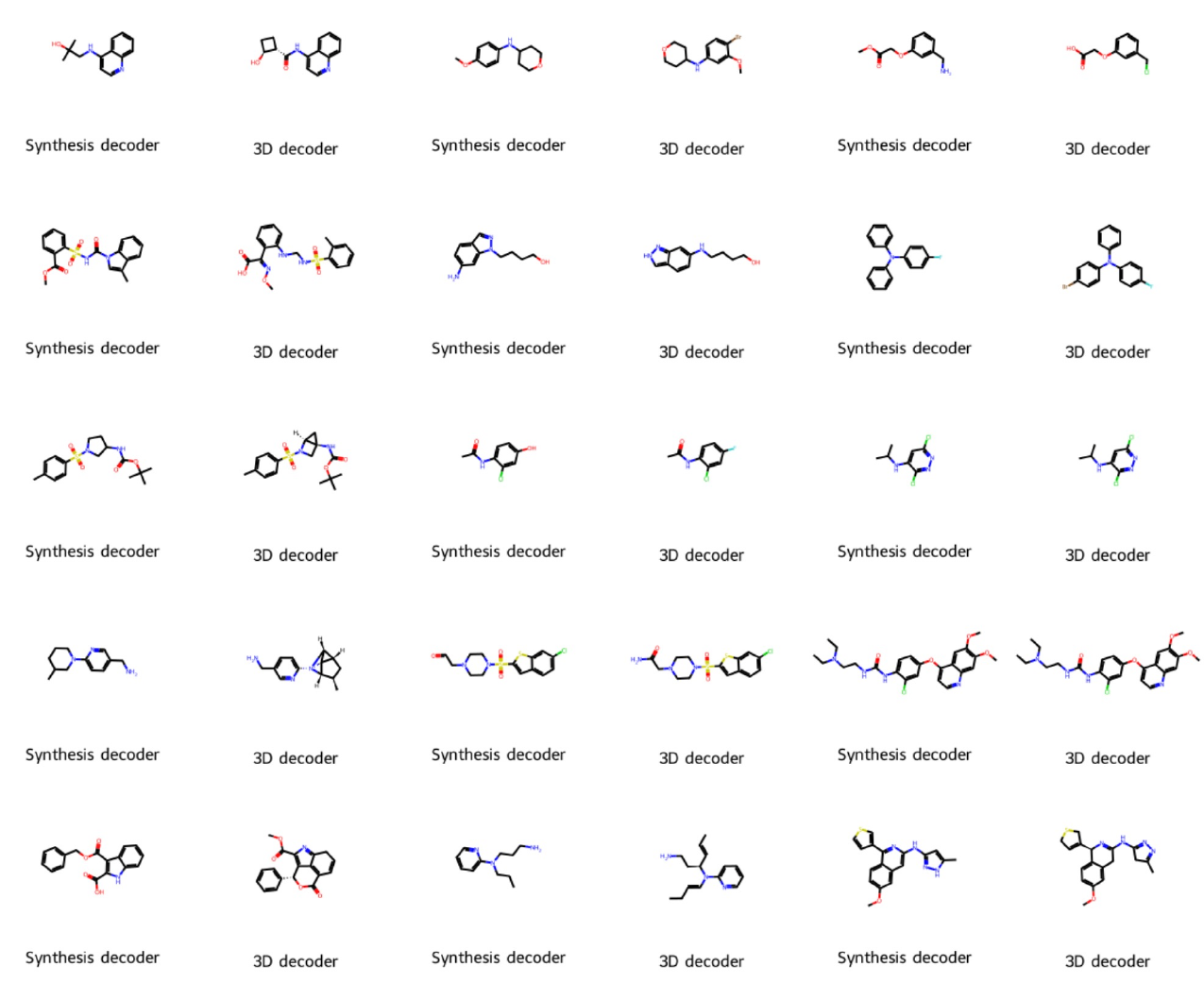}
    \caption{\textbf{Unconditional generation: synthesis- vs. 3D-decoded outputs.} Examples of decoded outputs from the synthesis decoder, followed by outputs of the 3D decoder, from the same latent.}
    \label{fig:app_syn_vs_3d_uncond}
\end{figure}

\subsection{Pharmacophore conditioned generation}
\label{sect:pharmCondGen}

During training, we load the pharmacophore features for a conformer and randomly drop $k$ features, where $k$ is sampled uniformly from $[0, \max(0,N_{\text{ph}}-3)]$, and $N_{\text{ph}}$ denotes the number of pharmacophore features in the sample. 
During sampling, we set the latent to contain $M$ atoms, with $M$ sampled uniformly from $[\max(1, N-3), N+3]$, where $N$ is the number of heavy atoms in the query (i.e., excluding hydrogens).

\begin{table}[htbp]
    \centering
    \footnotesize
    \caption{\textbf{Molecule quality metrics for in-distribution pharmacophore conditioning.} Higher is better for all.}
    \label{tab:app_id_pharmacophores}
    \begin{tabular}{@{}lccc@{}}
        \toprule
        \textbf{Method} &
        \multicolumn{1}{c}{\makecell[c]{\textbf{Validity}}} &
        \multicolumn{1}{c}{\makecell[c]{\textbf{Uniqueness}}} &
        \multicolumn{1}{c}{\makecell[c]{\textbf{Diversity}}} \\
        \midrule
        \ours (3D out)   & 0.78 & 0.60 & 0.72 \\
        \ours (syn out)  & 1.0 & 0.63 & 0.75 \\
        \bottomrule
    \end{tabular}
\end{table}

\begin{figure*}
    \centering
    \includegraphics[width=0.55\linewidth]{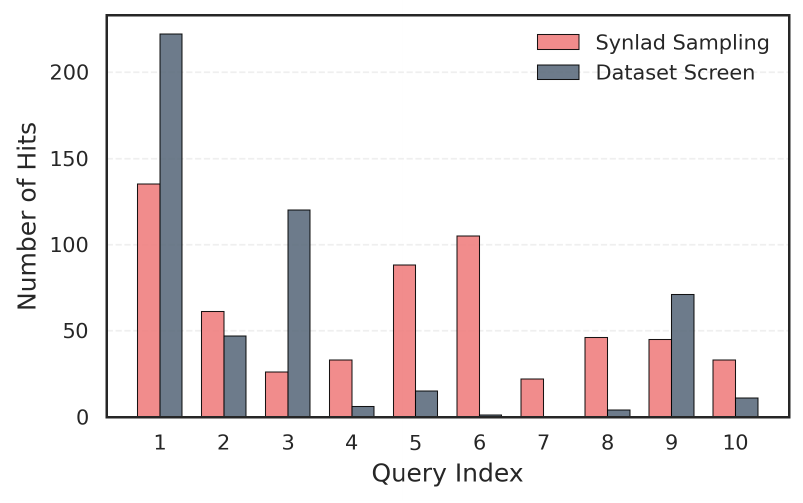}
    \caption{\textbf{Dataset screen vs. \ours sampling}. We report the number of unique hits for randomly selected queries from the test set. See the ``Screening case study'' section in the main text for further details.}
    \label{app:fig_screen_vs_synlad}
\end{figure*}

\subsection{Bioactive hit diversification experiment}
\label{app:pdb_exp}

\begin{figure}
    \centering
    \includegraphics[width=\linewidth]{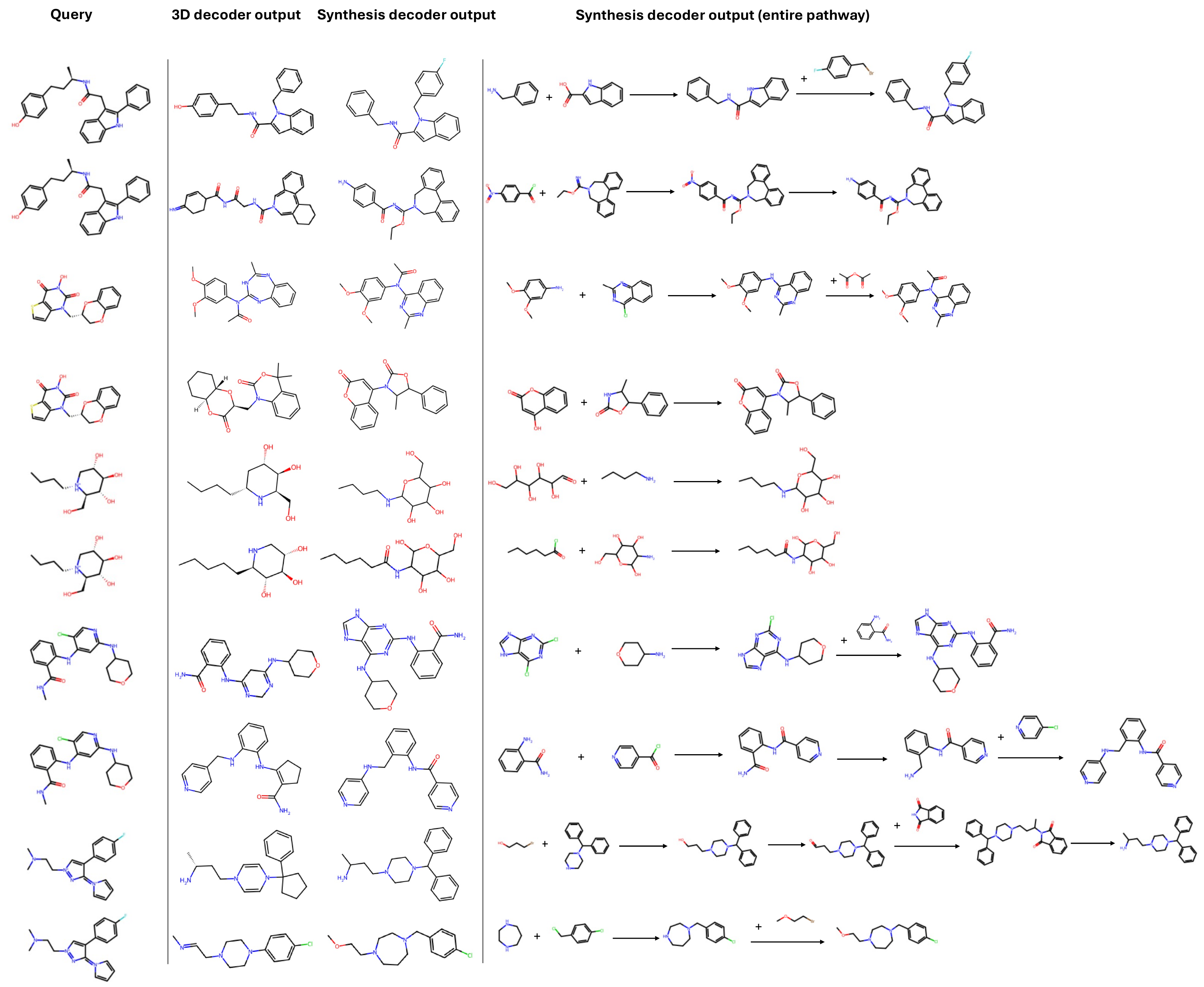}
    \caption{\textbf{\ours outputs for the Lit-PCBA conditioning task.} We visualize the Lit-PCBA ligand used as the conditioning \textbf{Query} (via its pharmacophore), together with the corresponding \textbf{3D decoder output} and the \textbf{synthesis decoder output} (final product and \textbf{entire pathway}). For readability, the query and 3D-decoded molecules are shown in 2D, although both are modeled in 3D coordinate space. We intentionally select examples where the synthesis- and 3D-decoder outputs do not coincide.}
    \label{fig:app_examples_litpcba}
\end{figure}

\begin{figure}
    \centering
    \includegraphics[width=0.6\linewidth]{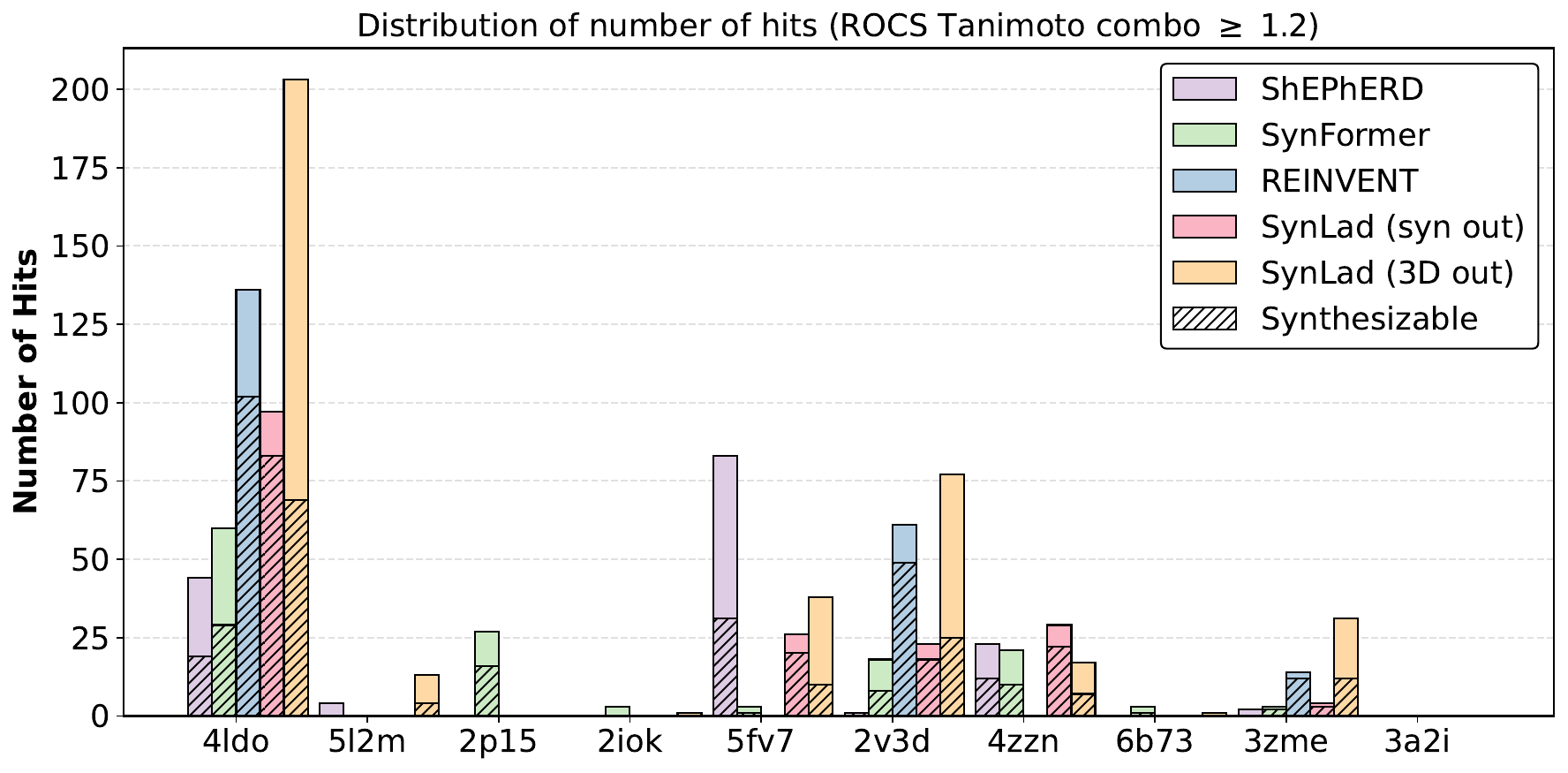}
    \caption{\textbf{Bioactive hit difersification experiment.} We add to the results in Figure~\ref{fig:pharm_cond_litpcba} the performance of the 3D decoder.}
    \label{fig:app_pharm_cond_litpcba_extra}
\end{figure}

We used ten PDB targets from \citet{litpcba}, with PDB IDs 4LDO, 5L2M, 2P15, 2IOK, 5FV7, 2V3D, 4ZZN, 6B73, 3ZME, 3A2I. We extracted the corresponding ligand for each of these as the `query'. We show examples of decoded outputs, together with the conditioning query in Figure~\ref{fig:app_examples_litpcba}. We show additional results (including the per-query number of hits for the 3D decoder) in Figure~\ref{fig:app_pharm_cond_litpcba_extra}.

\begin{figure}
    \centering
    \includegraphics[width=0.9\linewidth]{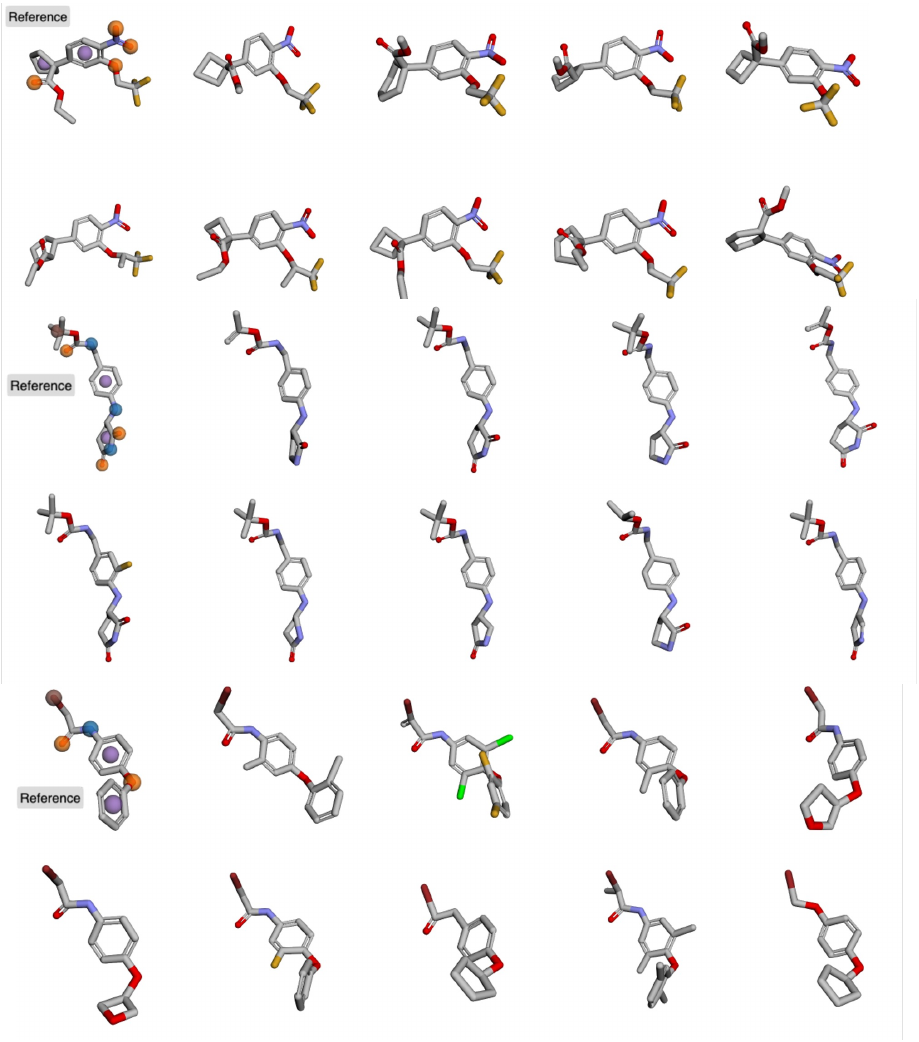}
    \caption{\textbf{Examples of query (reference) and sampled molecules conditioned on pharmacophores.} Samples were decoded using the 3D decoder, and we show the molecules in their predicted conformation.}
    \label{fig:ph4_3d_examples}
\end{figure}

\begin{figure}
    \centering
    \includegraphics[width=0.6\linewidth]{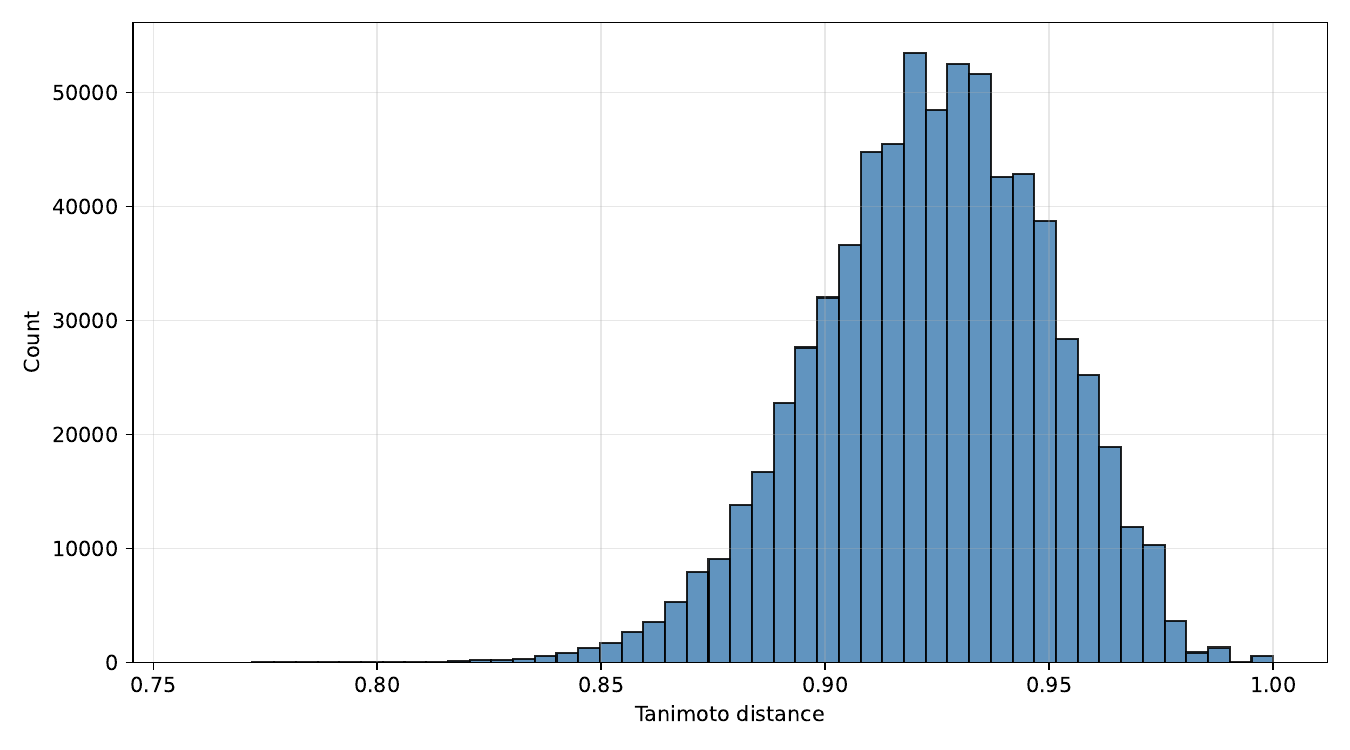}
    \caption{\textbf{Bioactive hit difersification experiment.} Tanimoto distances of Lit-PCBA ligands to our training set.}
    \label{fig:ood}
\end{figure}

For the hit diversification experiment, we quantify the extent to which molecules are out-of-distribution from the training set by plotting Tanimoto distances of the Lit-PCBA ligands to all training molecules in Figure \ref{fig:ood} (computed using 2048-bit Morgan fingerprints).

In addition to the results in Table~\ref{tab:pharm_cond_results}, we provide a ranking table that summarizes the relative performance of each method across all evaluated metrics. For each metric, methods are ranked per task and the average rank is reported, using a joint ranking scheme. This view highlights the consistency of each method across objectives, rather than emphasizing performance on any single metric. As shown in Table~\ref{tab:ranking_table}, \ours achieves the best or tied-best average rank for validity, unique scaffold hits, and synthesizable hits.

\begin{table}[h]
\centering
\caption{\textbf{Ranking table for hit diversification (OOD) task.} We report mean ranks using a joint ranking scheme. Lower is better.}
\label{tab:ranking_table}
\begin{tabular}{lccccc}
\toprule
\textbf{Method} & 
\textbf{Validity $\downarrow$} & 
\textbf{Hits (avg.) $\downarrow$} & 
\textbf{Unique scaff. hits (avg.) $\downarrow$} & 
\textbf{Max score $\downarrow$} & 
\textbf{Synthesizable hits $\downarrow$} \\
\midrule
SynLaD    & \textbf{1.10} & \textbf{2.10} & \textbf{1.90} & 2.20 & \textbf{2.00} \\
SynFormer & \textbf{1.10} & 2.20          & \textbf{1.90} & 2.00         & 2.10 \\
REINVENT  & 1.20          & \textbf{2.10} & 3.00          & \textbf{1.70} & 2.10 \\
ShePHERD  & 4.00          & 2.60          & 2.20          & 3.10      & 2.30 \\
\bottomrule
\end{tabular}
\end{table}

\FloatBarrier

\begin{figure}[h]
    \centering 
    \includegraphics[width=0.9\linewidth]{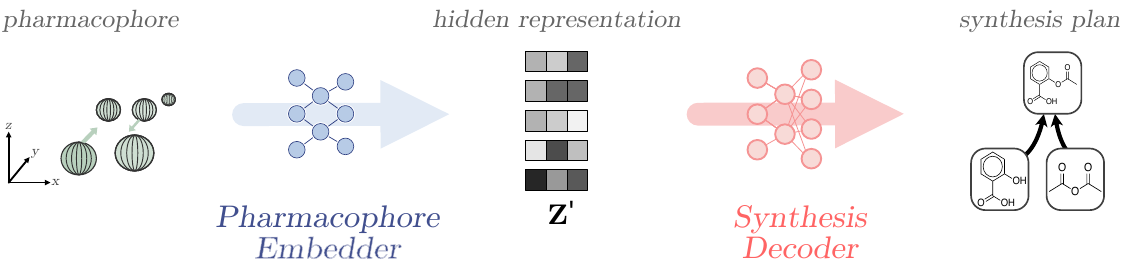}
    \caption{\textbf{Removing the 3D VAE and diffusion module.} To assess the importance of our model's autoencoder structure and 3D components, we run an ablation where we remove these. 
    Here, as shown, we instead train a model to  map directly from a pharmacophore to a synthesis plan via a deterministic hidden representation, $\mathbf{Z'}$. 
    For consistency with \ours, we keep the pharmacophore embedding network the same as that used as part of the conditional diffusion transformer, and we also ensure that the hidden representation $\mathbf{Z'}$ has the same dimensionality as the previously used stochastic latent variable, $\mathbf{Z}$. 
    The synthesis decoder also decodes from $\mathbf{Z'}$ in the same way as before, using cross-attention (see Figure~\ref{fig:modelOverview}, C).  
    }
    \label{fig:No3DAblationSchematic}
\end{figure}

\subsection{Ablations}
\label{app:ablation}

\begin{figure}[h]
    \centering
    \includegraphics[width=0.7\linewidth]{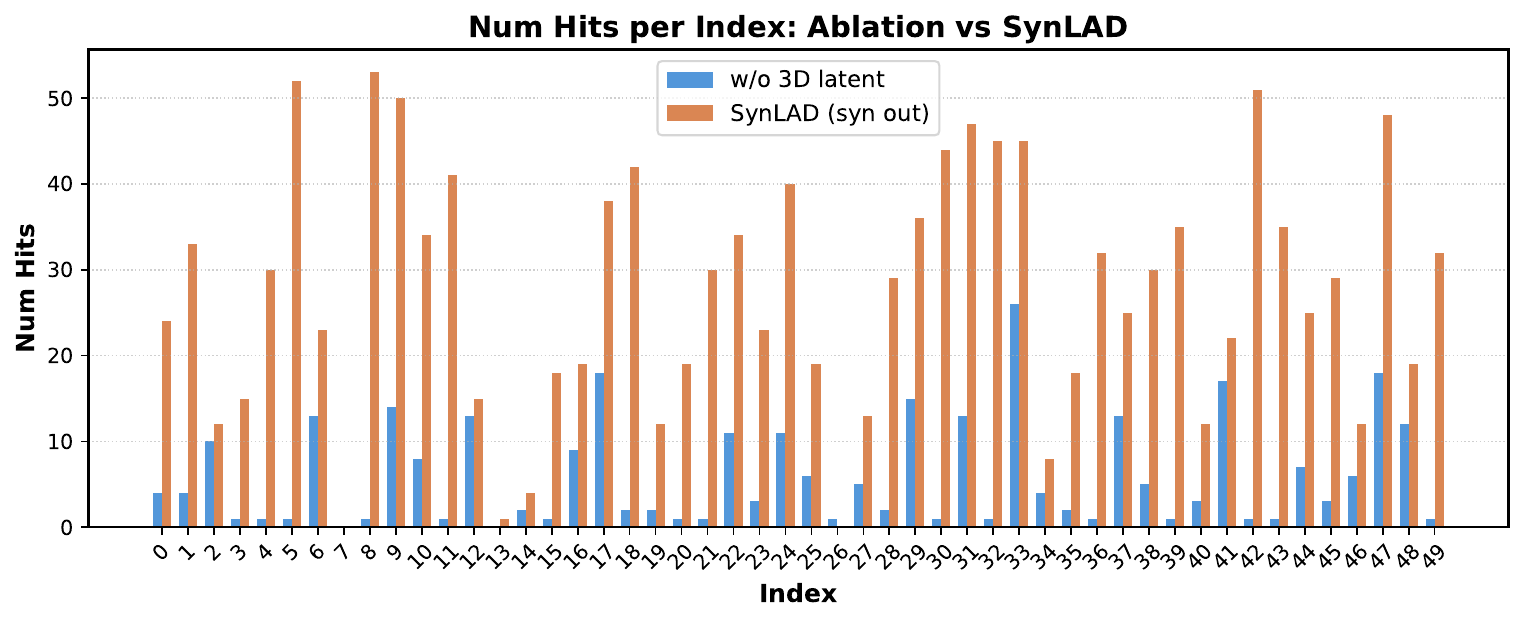}
    \caption{\textbf{Effect of removing the 3D branch on the number of hits.} We generate molecules conditioned on a pharmacophore with 1) \ours and a 2) synthesis decoder directly conditioned on the pharmacophore embedding.}
    \label{fig:app_ablation}
\end{figure}

\paragraph{Removing the 3D VAE and diffusion module.} In principle one could condition a synthesis decoder like the one in \ours, or equivalently others from the literature \citep{gao2024generative, lee2025rethinking} on pharmacophores, to have the model learn to generate synthesizable molecules that preserve those pharmacophores. To see whether there is benefit in modeling the conditioning input to the synthesis decoder using a latent diffusion framework as we do in \ours, we perform an ablation where we remove the 3D latent component from \ours and condition the synthesis decoder directly on the pharmacophore. We preserve the pharmacophore embedder described in Section~\ref{section:conditioning_method} and use the resulting embedding to condition on with an otherwise identical synthesis decoder via cross-attention. This yields a simplified pipeline, pharmacophore $\rightarrow$ pharmacophore embedding $\rightarrow$ synthesis pathway (see Figure~\ref{fig:No3DAblationSchematic}), which isolates the contribution of the 3D branch while keeping the synthesis decoder architecture and training procedure unchanged. We run the ablation on the in-distribution (ID) and out-of-distribution (OOD) conditioning tasks reported in Section~\ref{sect:experiments} and show in Table~\ref{tab:ablation_div_hits} that although the Tanimoto Combo scores do not suffer a drastic drop, the ablated version of the model appears to have collapsed to generating few samples per input (see Figure~\ref{fig:app_ablation}), showing that the entire framework is crucial to generating diverse hits.

\paragraph{Using a SMILES VAE.} We further investigate the effects of learning a conditional diffusion model for phramacophore-conditioned generation in a less expressive latent space. We replace the 3D autoencoder with a transformer VAE that operates on SMILES strings, to remove the 3D-reasoning component of \ours, while keeping the latent-conditioning component. SMILES strings are tokenized at the atom level using a regular-expression tokenizer adapted from \citet{schwaller2018found}. The SMILES VAE preserves the encoder-decoder architecture of the 3D autoencoder, and the same per-token diagonal-Gaussian latent. $\mathcal{L}_{3D}$ in Eq.~\eqref{eq:loss} is replaced by a cross-entropy loss over tokens, and $\lambda$ is set to 0.5. We keep an identical synthesis decoder, and train a second-stage latent diffusion model conditioned on pharmacophore embeddings. We run the ablation on both the ID and OOD conditioning tasks in Section~\ref{sect:experiments} and report results in Table~\ref{tab:ablation_div_hits}. We note that the diversity and uniqueness advantages of SynLaD are preserved, supporting our hypothesis that learning a pharmacophore-conditional distribution over latents via the diffusion module helps, promoting variability in the synthesis decoder’s conditioning. In contrast, the number of hits and Tanimoto scores decrease, suggesting that 3D information helps better capture the latent space for the pharmacophore-conditioning task.

\begin{table}
  \centering
  \footnotesize
  \caption{\textbf{Ablations.} We report metrics for our in-distribution (ID) and hit diversification / out-of-distribution (OOD) experiments. \textit{\ours (syn out)} refers to synthesis output of the baseline \ours model, \textit{w/o 3D VAE \& DM} refers to an ablation where we remove the 3D heads and diffusion module and condition the synthesis decoder directly on pharmacophore embeddings, and \textit{w/ SMILES VAE} refers to a model where we switch the 3D VAE for a SMILES VAE. Metrics are \textbf{averages} over all conditioning inputs.}
  \setlength{\tabcolsep}{3.2pt}
  \renewcommand{\arraystretch}{1.1}
  \begin{tabular}{llccccc}
    \toprule
    \textbf{{Setting}} &
    \textbf{Method} &
    \textbf{Div.} $\uparrow$ &
    \textbf{Uniq.} $\uparrow$ &
    \makecell[c]{\textbf{Tanimoto}\\\textbf{Combo} $\uparrow$} &
    \textbf{\#Hits} $\uparrow$ &
    \makecell[c]{\textbf{\#Scaff.}\\\textbf{hits} $\uparrow$} \\
    \midrule
    {ID} & \ours~(syn out) & 0.75 & 0.63 & 1.18 & 27.5 & 11.6 \\
    {ID} & w/o 3D VAE \& DM   & 0.49 & 0.17 & 1.12 & 5.9  & 3.5 \\
    {ID} & {w/ SMILES VAE}   & {0.72} & {0.55} & {1.07} & {20.7}  & {8.9} \\
    \midrule
    {OOD} & {SynLaD~(syn out)} & {0.86} & {0.79} & {1.29} & {17.9} & {6.9} \\
    {OOD} & {w/o 3D VAE \& DM}   & {0.76} & {0.18} & {1.11} & {4.4}  & {1.5} \\
    {ID} &{ w/ SMILES VAE}   & {0.86} & {0.77} & {1.19} & {10.4}  & {3.2} \\
    
    \bottomrule
  \end{tabular}
  \label{tab:ablation_div_hits}
\end{table}

\begin{figure}
    \centering
    \includegraphics[width=0.7\linewidth]{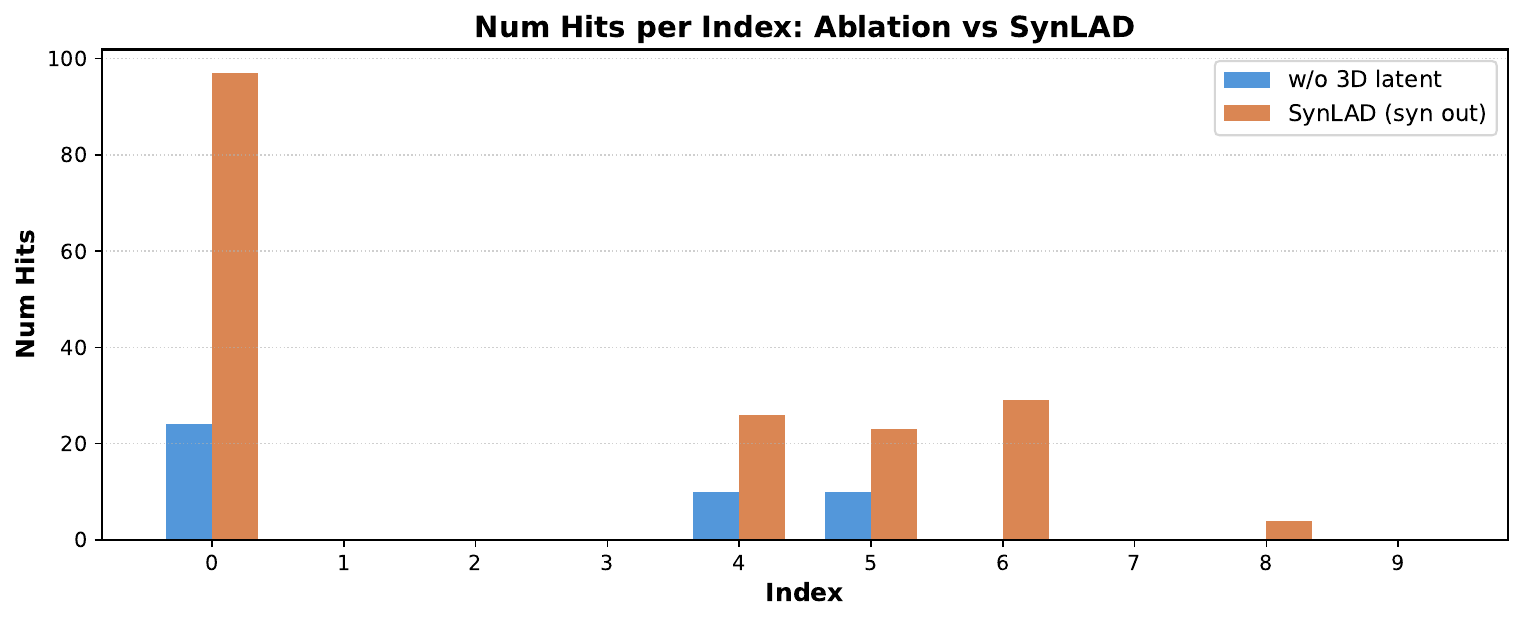}
    \caption{\textbf{Effect of removing the 3D branch on the number of hits, for our OOD hit diversification experiment.}} We replicate the setting in Figure \ref{fig:app_ablation}.
    \label{fig:app_ablation2}
\end{figure}

\subsection{Comparison to baselines}
\label{app:baselines}

\paragraph{ShEPhERD} We used the official implementation at \url{https://github.com/coleygroup/shepherd} \citep{adamsShEPhERD2024} and use the $p(x_1|x_3, x_4)$ conditional setting provided in the repository where $x_1$ denotes molecular structure, $x_3$ is the query electrostatic potential surface, and $x_4$ is the query's pharmacophore. We generate 500 analogues with number of atoms uniformly sampled from the interval $[\max(3, N-6), N+6]$, where $N$ is the number of atoms in the query molecule (including hydrogens), to best match \ours's settings. 

\paragraph{SynFormer} We used the official implementation at \url{https://github.com/wenhao-gao/synformer} and changed the following inference settings to allow for higher quality designs compared to the default: \texttt{search_width=32, exhaustiveness=128, time_limit=300}. We generate and evaluate 500 molecules per query.

\paragraph{REINVENT}
We used the official implementation of REINVENT 4 available at \url{https://github.com/MolecularAI/REINVENT4} \citep{loeffler2024reinvent-f98,Blaschke2020,olivecrona2017molecular-7b7}. We start from the default REINVENT prior (from \url{https://zenodo.org/records/15641297}), which was trained on ChEMBL \citep{mendez2019chembl-4bd} to cover a broad region of chemical space. 
We followed the repository's example reinforcement learning configuration (including hyperparameters), using a Tanimoto color scorer (with up to 5 conformers for computational speed) and a maximum of 300 steps (we do not include any other scoring components). We use the final 500 sampled molecules for evaluation.
Although additional steps and hyperparameter tuning may further improve results, each target run already takes approximately 6 hours and can involve scoring up to \num{30000} proposed molecules; therefore, we defer a comprehensive hyperparameter study (for both REINVENT and \ours) to future work. 

\end{document}